\documentclass[11pt]{article}

\usepackage{microtype}
\usepackage{graphicx}
\usepackage{subfigure}
\usepackage{booktabs}
\usepackage{hyperref}
\usepackage{authblk}
\usepackage{caption}
\usepackage[most]{tcolorbox}
\usepackage{multirow}

\usepackage{amsmath,amssymb,mathtools,amsthm}
\usepackage[capitalize,noabbrev]{cleveref}
\usepackage[textsize=tiny]{todonotes}
\usepackage{tikz}

\newtheorem{example}{Example}[section]
\usepackage[margin=1in]{geometry}

\usepackage{etoolbox} 

\newcommand{\authorblock}[4]{%
  {\large\bfseries #1\par}%
  \vspace{0.2em}%
  {#2\par}%
  {#3\par}%
  \vspace{0.2em}%
  {\ttfamily #4\par}%
}

\usepackage{authblk}
\author{} 

\title{\bf Evaluating AI Grading on Real-World Handwritten College Mathematics:\\
A Large-Scale Study Toward a Benchmark}

\date{} 

\begin{document}

\maketitle

\vspace{-4.0em} 

\begin{center}
\begin{minipage}[t]{0.32\textwidth}\centering
\authorblock{Zhiqi Yu}{UC Irvine}{Irvine, USA}{zhiqiy10@uci.edu}
\end{minipage}\hfill
\begin{minipage}[t]{0.32\textwidth}\centering
\authorblock{Xingping Liu}{Independent}{Shanghai, China}{xpliu203@gmail.com}
\end{minipage}\hfill
\begin{minipage}[t]{0.32\textwidth}\centering
\authorblock{Haobin Mao}{UC Berkeley}{Berkeley, USA}{haobinmao@berkeley.edu}
\end{minipage}

\vspace{1.3em}

\begin{minipage}[t]{0.32\textwidth}\centering
\authorblock{Mingshuo Liu}{UC Irvine}{Irvine, USA}{mingsh15@uci.edu}
\end{minipage}\hfill
\begin{minipage}[t]{0.32\textwidth}\centering
\authorblock{Long Chen}{UC Irvine}{Irvine, USA}{lchen7@uci.edu}
\end{minipage}\hfill
\begin{minipage}[t]{0.32\textwidth}\centering
\authorblock{Jack Xin}{UC Irvine}{Irvine, USA}{jack.xin@uci.edu}
\end{minipage}

\vspace{1.3em}

\begin{minipage}[t]{0.6\textwidth}\centering
\authorblock{Yifeng Yu$^{\star}$}{UC Irvine}{Irvine, USA}{yifengy@uci.edu}
\end{minipage}

\vspace{0.8em}

\end{center}


\renewcommand{\thefootnote}{\fnsymbol{footnote}}
\footnotetext[1]{Corresponding author.}
\renewcommand{\thefootnote}{\arabic{footnote}}
\setcounter{footnote}{0}

\vspace{1.0em} 

\noindent\textbf{Keywords:} AI grading; OCR + large language models; mathematical reasoning; rubric-guided evaluation; formative feedback; real-world benchmarks; human–AI agreement.

\begin{abstract}
Grading in large undergraduate STEM courses often yields minimal feedback due to heavy instructional workloads. We present a large-scale empirical study of AI grading on real, handwritten single-variable calculus work from UC Irvine. Using OCR-conditioned large language models with structured, rubric-guided prompting, our system produces scores and formative feedback for thousands of free-response quiz submissions from nearly 800 students. In a setting with no single ground-truth label, we evaluate performance against official teaching-assistant grades, student surveys, and independent human review, finding strong alignment with TA scoring and a large majority of AI-generated feedback rated as correct or acceptable across quizzes.

Beyond calculus, this setting highlights core challenges in OCR-conditioned mathematical reasoning and partial-credit assessment. We analyze key failure modes, propose practical rubric- and prompt-design principles, and introduce a multi-perspective evaluation protocol for reliable, real-course deployment. Building on the dataset and evaluation framework developed here, we outline a standardized benchmark for AI grading of handwritten mathematics to support reproducible comparison and future research.
\end{abstract}

\section{Introduction and Motivation}
Grading at scale remains one of the most persistent challenges in undergraduate education. In high-enrollment STEM courses such as calculus, teaching assistants (TAs) face thousands of open-response submissions under tight deadlines. The resulting workload often leads to two systemic problems: (1) declining grading quality and consistency, and (2) minimal or no personalized feedback—although meaningful and informative feedback is essential for learning. These issues directly undermine fairness, transparency, and student engagement in foundational mathematics instruction. Since calculus is one of the largest and most resource-intensive gateway courses in STEM education nationwide, scalable grading solutions have potential impact across thousands of sections and tens of thousands of students each year.

To address this, we developed and deployed a large-scale AI grading system that combines optical character recognition (OCR) with large language models (LLMs) to assess handwritten solutions and generate rubric-aligned feedback. AI-generated scores were provided for reference only; TA scores remained the official course grades. To our knowledge, this is among the first large-scale real-course studies of LLM-based grading in university mathematics.

We deployed the system across three academic terms in a large-enrollment single-variable calculus sequence at UC Irvine, covering 20 quizzes and more than 1,000 students. The sequence consists of two consecutive courses Math 2A and Math 2B.

\begin{enumerate}
\item \textbf{Winter 2025.} We tested the system on two quizzes in Math 2B, each containing three free-response questions, in a class of over 500 students.

\item \textbf{Spring 2025.} We applied the system to six quizzes in Math 2A (two questions each, one free-response and one multiple-choice or fill-in-the-blank, approximately 400 students) and six quizzes in Math 2B (two or three free-response questions, approximately 360 students). The Spring 2025 datasets served as the primary basis for system refinement; five quizzes from this term were used for detailed comparison against TA scores, student surveys, and independent human reviewers.

\item \textbf{Fall 2025.} We further validated the system on six additional quizzes in Math 2B, each containing two or three free-response questions, with approximately 200 students.
\end{enumerate}

All quantitative evaluations in this paper focus exclusively on free-response questions from 2025 Spring, as multiple-choice and fill-in-the-blank items do not reflect the core challenges of AI-based handwriting interpretation and rubric-guided partial-credit assessment and the production of clear, constructive feedback.

We evaluate AI-assisted grading via TA-score alignment, student surveys, and an independent human  review (over twenty reviewers). AI-assigned scores show strong agreement with human grading, and the generated feedback is generally judged correct or acceptable by independent reviewers, with remaining errors concentrated in OCR failures and a small set of rubric edge cases. Together, these results suggest that the system can reduce TA workload while providing more detailed feedback to students; See Section \ref{sec:humanreview} for full quantitative results.

From a technical standpoint, one crucial aspect of grading handwritten mathematics with LLMs is prompt and rubric design. Because there is no unique ground-truth score for many partially correct solutions, the system must translate free-form handwritten student work into consistent, fine-grained judgments that remain aligned with course rubrics. Beyond coping with noisy OCR and varied handwriting, we therefore develop and iteratively refine a structured prompt framework that combines system-level instructions, problem-specific reference solutions, and machine-readable rubrics. This framework, described in Section~\ref{sec:prompt}, is crucial to stabilizing scores and generating useful feedback across thousands of heterogeneous submissions.

Recent work \cite{aisyah20,mok2024,caraeni2025,lee2025,lee2025mathvaluator,liu2024,nkoyo2025},etc on AI-assisted grading of handwritten STEM solutions has grown rapidly with recent LLM advances, but typically emphasizes score reliability rather than course-scale evaluation of feedback quality. We study course-scale deployment on thousands of authentic handwritten calculus submissions and evaluate both scores and feedback using TA grades, student surveys, and independent human review (20+ reviewers).

\subsection*{Main Contributions} Our focus is on an underexplored ML problem: how to reliably evaluate and deploy LLM-based reasoning systems without ground truth, under noisy inputs, partial-credit grading, and high-stakes constraints.
\begin{itemize}
    \item \textbf{Large-scale, real-world deployment.}
    We report a large-scale deployment of an end-to-end OCR+LLM grading system (scores and feedback) on free-form handwritten calculus solutions in a high-enrollment, for-credit university course, using thousands of authentic quiz submissions.

    \item \textbf{Prompt-and-rubric grading pipeline.}
    We develop a practical pipeline that couples OCR-conditioned reasoning with carefully designed system prompts and a multi-rubric strategy (typically a flexible and a fixed rubric). The prompts and rubrics are iteratively refined using failure-case analysis to improve partial-credit scoring and formative feedback.

    \item \textbf{Multi-perspective human-in-the-loop evaluation.}
    We introduce a structured evaluation protocol combining TA score alignment, student feedback, and independent reviews from over twenty external reviewers to assess both score reliability and the accuracy/usefulness of AI-generated feedback, where no ground-truth labels exist.

    \item \textbf{Foundation for a future benchmark.}
We have assembled a large collection of anonymized handwritten solutions together with OCR outputs, tuned rubrics, AI-generated scores and feedback, and independent human review results. As future work, we plan to release benchmark subsets in two tracks. \emph{Track A (clean)} will pair scanned work with corrected transcriptions and instructor/TA--approved reference grades, enabling controlled evaluation of OCR and rubric-based grading. \emph{Track B (noisy)} will retain the original OCR outputs and reviewer labels to evaluate robustness and triage---i.e., when the system should defer to human review under noisy inputs, and how reliably it can flag likely OCR-induced errors.
\end{itemize}

Although this work focuses on single-variable calculus, the challenges it addresses—scalability, fairness, and feedback quality—are shared across many large STEM courses, positioning foundational mathematics as a natural testbed for broader AI-assisted assessment.

\section{Overview of the AI Grading Pipeline}


\begin{figure*}
    \centering
    \includegraphics[width=0.8\linewidth]{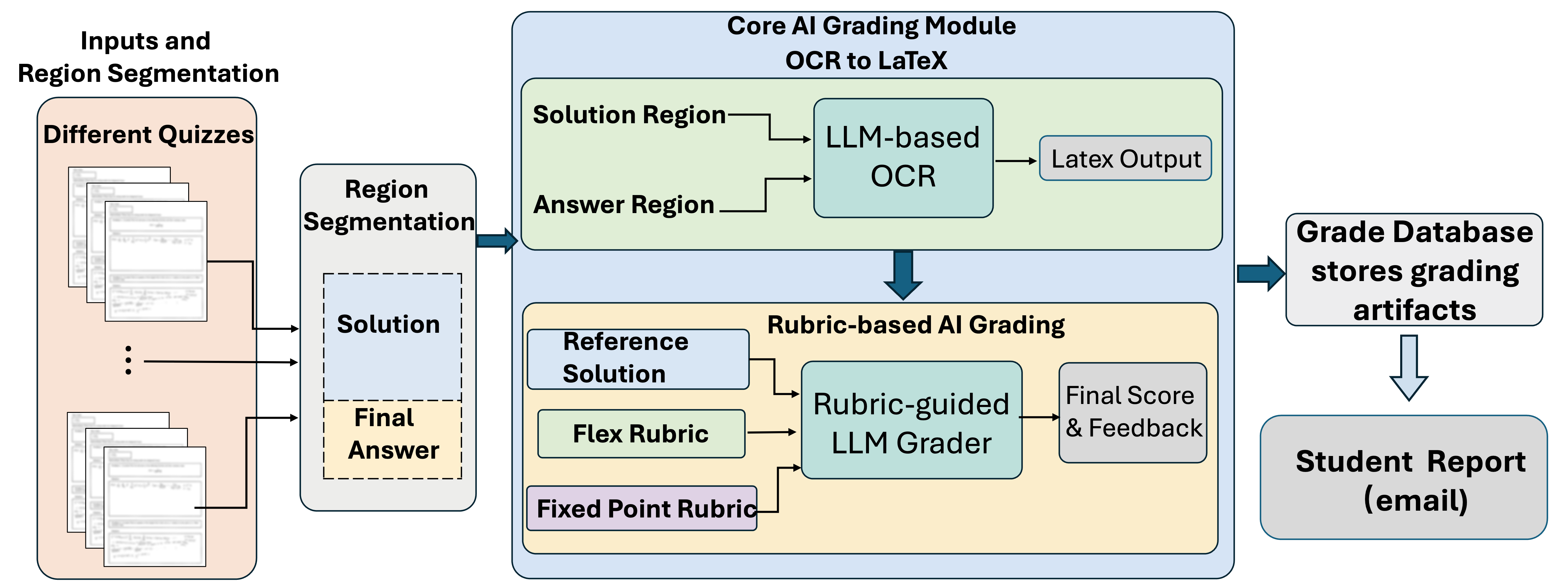}
    \caption{Overview of the proposed AI grading pipeline: The core module converts handwritten solutions to LaTeX and evaluates them against reference solutions and structured rubrics, producing detailed scores and feedback.}
    \label{fig:pipeline}
\end{figure*}

 Our pipeline consists of three modular stages (Figure~\ref{fig:pipeline}): region segmentation, OCR-to-\LaTeX{} transcription, and rubric-based LLM grading with feedback. We use a standardized answer-sheet format  that separates each problem into a solution region and a final-answer region, enabling the pipeline to handle variable numbers of problems by processing each region independently. See Appendix Section~\ref{section:answersheet}. We perform OCR and grading at the region level (rather than on whole-page inputs) to reduce cross-problem interference and hallucination; free-form solution regions remain the most challenging. The resulting scores and feedback are delivered to students by email. See Figure \ref{fig:pipeline}.

\textbf{1. OCR and Transcription.}
Handwritten student work is digitized via automated OCR and converted to \LaTeX\ to preserve mathematical structure. Among the tools we evaluated, ChatGPT~4.1~mini paired with well-designed prompts achieved a favorable trade-off among accuracy, scalability, and cost; see Section~\ref{sec:ocr-eval} for details.

\textbf{2. Rubric-Guided Prompting.}
We use a two-level prompt design: a system message that provides global grading principles and output constraints, and problem-specific rubrics encoded as structured prompts specifying scoring and feedback criteria, output format, and exemplar solutions. Prompt templates are iteratively refined on sample submissions to better align LLM outputs with rubric intent (Section~\ref{sec:prompt}).

\textbf{3. LLM Evaluation.}
The transcribed solutions and rubric prompts are processed by large language models, primarily GPT-4.1-mini, with supplementary evaluations using o3-mini if needed, which produce both numerical scores and explanatory feedback. Because grading lacks an absolute ground truth, we assess reliability via three complementary validations: alignment with TA scores, student feedback, and independent human reviews (Section~\ref{sec:humanreview}).

\paragraph{Model choice and evolving capabilities.}
Unless otherwise noted, we use GPT-4.1-mini for both OCR and grading, balancing accuracy, latency, and cost for course-scale deployment. We additionally use o3-mini for a small subset of algebra-heavy or otherwise challenging cases where stronger reasoning is needed. See section~\ref{sec:o3-vs-4.1} for comparisons. Rather than chasing each new model release, we keep the pipeline stable and allow the backend model to be swapped as capabilities improve; systematic cross-model comparisons and routing policies are left to future work.

\section{OCR Component}\label{sec:ocr-eval}

For transcribing handwritten mathematical work, we evaluated several OCR options on our course data in Fall 2024/Winter 2025 and selected Mathpix as the most reliable choice at the time. After the release of GPT-4.1/4.1-mini in April 2025, we re-ran our OCR evaluation and found that, with OCR-specific prompting, GPT-4.1-mini produces higher-quality transcriptions on our data. Across thousands of handwritten submissions, around 88\% of transcriptions are \emph{acceptable}, meaning they are either (i) \emph{completely correct} (character-level \LaTeX\ fidelity) or (ii) \emph{sufficient for rubric-based grading}. Unlike Mathpix, which primarily targets literal transcription, GPT-4.1-mini can leverage contextual reasoning to resolve ambiguities, improving robustness to messy and heterogeneous student handwriting. In a head-to-head comparison on a fixed challenging subset (the {\bf{171-solution subset}}), consisting of 59 quizzes, each containing the same three questions (171 solutions in total), GPT-4.1-mini achieved 84\% acceptable transcriptions, compared to 55\% for Mathpix. See Appendix Section~\ref{section:gptvsmathpix} for representative examples.

OCR performance with GPT-4.1-mini is sensitive to prompt design; we use context-aware constraints to improve transcription stability. Below is an example of our basic OCR prompt.

\bigskip

\resizebox{0.9\linewidth}{!}{%
\begin{tcolorbox}[colback=gray!5,colframe=black,title=OCR prompts, fontupper=\small]
This problem asks the student to estimate a definite integral using a left Riemann sum:
\[
\int_1^5 (x^2 - x - 1)\,dx
\]
with **four subintervals** and **left endpoints**, without evaluating the exact limit.

\begin{enumerate}
    \item Use OCR to identify any text in this image and convert it into  LaTeX code
    \item  Do not correct any math or logical error or typos.  
\end{enumerate}
\end{tcolorbox}
}

\bigskip

To better utilize the contextual reasoning ability of GPT-4.1-mini OCR, it is important to state the problem in the prompts.  The other two targeted prompt strategies address two common, prompt-addressable sources of transcription instability.
\begin{itemize}
    \item \textbf{Hallucination.} When answer boxes are blank or sparsely written, GPT-4.1-mini may generate full solutions.  
    The above Prompt 1 and pre-printing the words “Solution’’ and “Final Answer’’ inside boxes stabilize the output and suppress hallucination (see Appendix Section~\ref{section:answersheet}).
    
   \item \textbf{Autocorrection.} GPT-4.1-mini occasionally corrects students’ mathematical mistakes during transcription. An explicit instruction not to correct errors (Prompt~2) helps reduce this behavior, though rare cases remain (less than 2\% of our samples). A representative example is provided in Appendix Section~\ref{sec:autocorrection}.

\end{itemize}

Two major OCR challenges remain: (1) geometric diagrams, where \texttt{GPT-4.1-mini} shows limited ability and, in our experiments, Mathpix did not reconstruct any diagrams from student work; and (2) reliably handling crossed-out or heavily scribbled text. In the latter case, OCR systems may transcribe erased work or drop nearby valid symbols when the crossed-out region visually merges with adjacent ink. We experimented with several prompt refinements to suppress erased work, but this often also removed nearby valid content.

Beyond these prompt-addressable issues, we analyzed all 27 problematic OCR cases within the \textbf{171-solution subset}. Most {GPT-4.1-mini} failures are layout-related, with the dominant failure mode involving complex division expressions. In particular, when nested fractions are written with insufficient spacing, the scope of the main fraction bar becomes ambiguous, causing terms to be merged, misparsed, or omitted. Other error types are uncommon. See Appendix~Section~\ref{app:4.1-ocr-fail} for an illustrative example of this fraction-related failure mode.

\section{Grading and Structured Prompt Engineering}\label{sec:prompt}

With OCR reliability largely established, a central challenge in AI grading lies in prompt design—specifically, constructing structured prompts that ensure consistency, fairness, and rubric alignment at scale. Our experiments show that careful system-message engineering and rubric formulation are decisive factors in achieving human-level grading accuracy.

\subsection{System Message Design}\label{sec:systemmessage}

Prompts in our grading system consist of two parts: a \textit{system message} that defines grading behavior and a \textit{rubric section} specifying problem-level criteria. We observed that API-based grading can be less thorough than grading performed in the interactive web interface. While the exact cause is not externally documented, this motivated us to make all grading instructions explicit in the system message so that API-based deployment achieves consistent, high-quality outputs.

To reduce stochasticity, we set temperature to near-zero values (0.0–0.1) for {GPT-4.1-mini} to improve response stability and reproducibility. For o3-mini, temperature is not an exposed API parameter and therefore cannot be adjusted. The system message was iteratively refined to emphasize logical consistency, factual correctness, and human-style feedback.
Below is an excerpt:

\bigskip

\begin{tcolorbox}[colback=gray!5,colframe=black,title=System message excerpt, fontupper=\small ]
{\it You are an expert math grader. When evaluating, please:
\begin{enumerate}
  \item Distinguish between mathematically correct statements and incorrect reasoning.  
  \item Avoid contradictions in feedback.  
  \item Do not penalize false starts if later corrected.  
  \item Provide concise, constructive explanations for point deductions.  
  \item Write feedback in clear, natural language suitable for students.
\end{enumerate}}
\end{tcolorbox}

\bigskip

These structured instructions help reduce logical inconsistency and excessive penalization of harmless early mistakes—common failure modes under naive prompting. Representative examples are provided in Appendix Section~\ref{app:systemmessage}.

\subsection{Rubric and Prompt Framework}

\begin{figure}[htbp]
  \centering
  \includegraphics[width=0.7\linewidth]{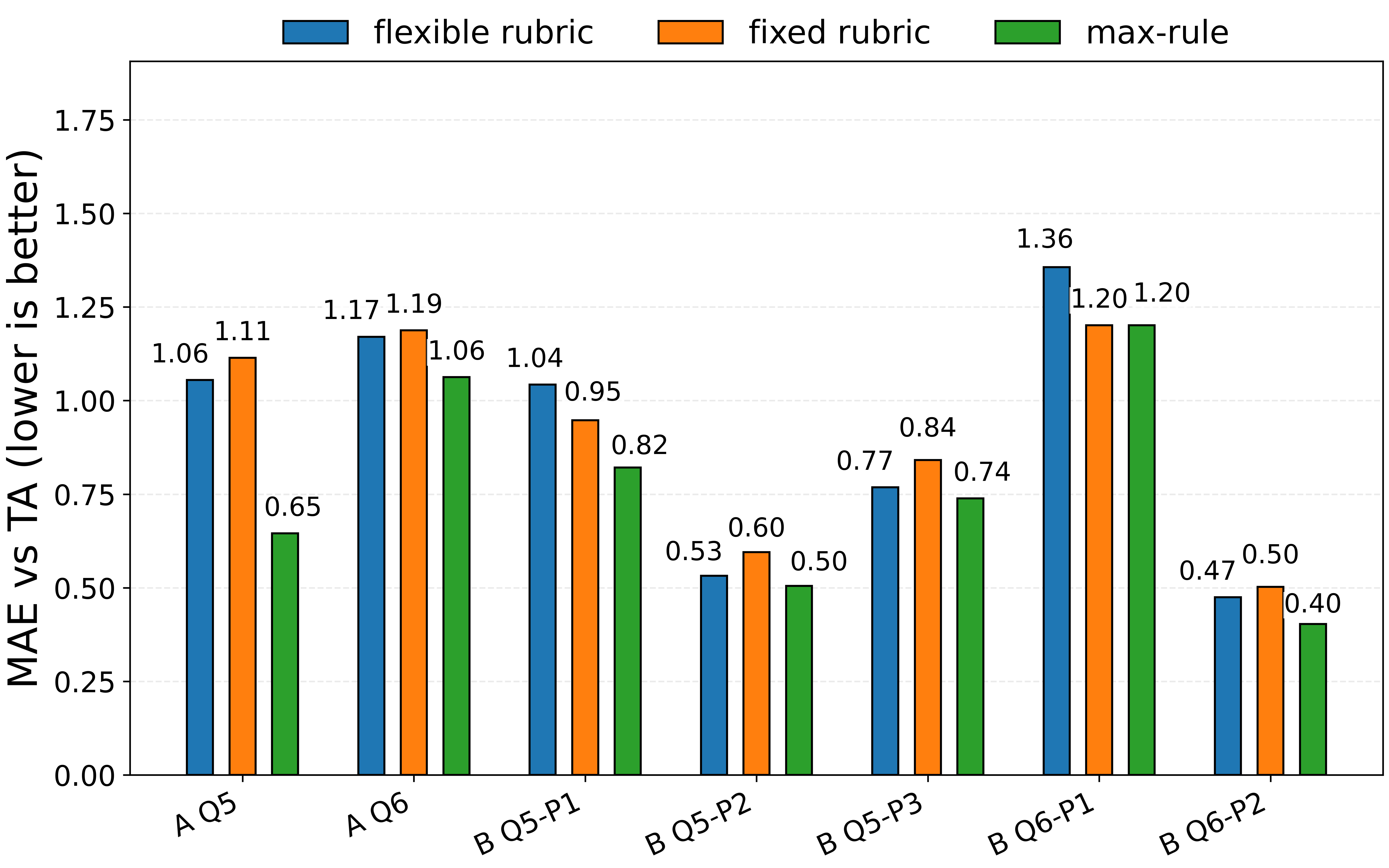}
  \caption{Bars correspond to the flexible rubric, fixed rubric and max-rule, respectively (left to right within each group): A-Math 2A, B-Math 2B}
  \label{fig:maxrule-mae}
\end{figure}

Our rubric framework, developed from thousands of Math 2A and B submissions, is designed to balance interpretability and grading precision. Each problem may be associated with multiple rubrics to accommodate different valid solution strategies. For most questions, we adopt a dual-rubric design consisting of:
\begin{enumerate}
  \item \textbf{Concise reference solutions}, summarizing one or more canonical correct approaches.
  \item \textbf{Two complementary rubrics}:
    \begin{itemize}
      \item \textbf{Flexible rubric} — human-like and tolerant of diverse reasoning styles; supports free-form evaluation to credit partially correct work and uncommon but valid approaches.
      \item \textbf{Fixed rubric} — fine-grained and point-based, with problem-specific checks targeting required steps and common errors in standard solutions.
    \end{itemize}
    See Appendix~\ref{app:dual-rubrics} for a representative example comparing the two rubrics and their JSON outputs, illustrating how the flexible rubric complements a checklist-style rubric on non-canonical but correct solutions. Importantly, flexibility does not systematically inflate scores: in our experiments, when locally correct steps are globally inconsistent or numerically incorrect, the flexible rubric may assign lower scores than the fixed rubric.
  \item \textbf{Selection rule.}
    To reduce model confusion, each submission is evaluated using both rubrics via independent API calls, and we adopt a \emph{max-rule}: the higher of the two rubric scores is selected as the final grade, together with the corresponding feedback. We evaluate this design on Math 2A and B quizzes~5--6, which contain more challenging problems and admit a wider range of valid solution paths and partial attempts.

    To assess alignment with human grading behavior, we compare grades produced by the fixed rubric alone, the flexible rubric alone, and the max-rule against TA scores (used as a reference rather than an absolute ground truth). Using mean absolute error (MAE) as the primary metric, the max-rule consistently achieves the lowest MAE across both courses and quizzes (Fig.~\ref{fig:maxrule-mae}). These results suggest that the max-rule yields more stable and human-aligned scores by crediting valid reasoning that may be under-recognized by a single rubric.
\end{enumerate}

\textbf{Exemplar-guided rubric generation.}
We use examples to help the model write new grading rubrics in a consistent format. Using primarily Spring~2025 Math 2A and B quizzes, we built a small library of representative problems by problem type (e.g., definite integrals) and refined a high-quality rubric for each. For a new problem in the same or a closely related category in a later term (e.g., Fall~2025), we provide the problem statement, a reference solution, and the closest existing rubric to a large language model, and ask it to produce a rubric at the same level of detail. The resulting rubric typically needs only light human editing (e.g., adding a newly observed misconception or a course-specific convention) before deployment. Future work will organize these rubrics into a lightweight database and automate more of this rubric-writing step.

\noindent\textbf{Two sources of potentially unjustified point deductions.}
We observed two recurring cases in which an LLM grader may deduct points even when the underlying mathematics is correct. 
First, the model may penalize disorganized or nonstandard presentation, despite correctly recognizing a valid solution. 
Second, our answer sheet separates the work area from the final-answer box: contextual reasoning improves OCR accuracy in the solution area, but the final-answer box provides little surrounding context and is therefore more prone to OCR errors. 
Consequently, the final answer may be misread (making a correct response appear incorrect) even when the solution transcription is accurate and the mathematics is valid. 

To mitigate these issues, we plan to incorporate a human-review fallback in future deployments when correct answers are clearly supported but full credit is not assigned. In the reported experiments, we already include rubric guidance that discourages style-based penalties and tolerates minor OCR-induced discrepancies in the final-answer box when the correct value is unambiguously supported by the solution steps. Representative rubric guidance used in our experiments is provided in Appendix Section~\ref{app:rubircguidance}.

\section{o3-mini vs. GPT-4.1-mini Grading Variance and Targeted Accuracy Evaluation}\label{sec:o3-vs-4.1}

We compare o3-mini and GPT-4.1-mini on grading accuracy and score variance using the algebra-heavy \textbf{171-solution set}. To isolate grader effects, we fix OCR to 4.1-mini and vary only the grading model (GPT-4.1-mini vs.\ o3-mini). As o3-mini emphasizes multi-step reasoning, we expect it to perform better on algebraically involved items; two independent reviewers found feedback similarly accurate overall but preferred o3-mini when differences arose. However, o3-mini has higher variance and cost, so we reserve it for the most challenging items and apply simple multi-run stabilization for reliability. See Appendix~\ref{app:4.1-o3-comparison} for details.

\noindent\textbf{Variance reduction.}
For each item, we run o3-mini three times, take the mean score, and output the run closest to this mean (with its feedback). Inspired by self-consistency in LLM reasoning as a stability signal \cite{wang2023}, we avoid majority voting—which typically requires many independent runs—and instead use this closest-to-the-mean rule for continuous partial-credit grading. Two independent human reviewers preferred the selected run in majority cases for feedback accuracy. See Appendix Section~\ref{app:an-shiyi} for a comparison with majority vote in  fluctuating cases.

\noindent\textbf{Human review.}
If exactly one of the three runs assigns full credit while the other two do not, we flag the item for human review.

\section{Evaluation Against Human Grading and Feedback}\label{sec:humanreview}
We evaluate AI-assisted grading for single-variable calculus in two courses (A and B), covering nearly 800 students, and assess both score alignment and feedback quality. A central challenge is the absence of a ground-truth benchmark: there is no publicly available expert-graded dataset for calculus solutions, and human graders themselves show nontrivial disagreement, especially on borderline or unconventional answers (Appendix~\ref{app:tworeviewerdifference}). This variability makes it difficult to define a universally accepted ``correct'' score, and it makes feedback evaluation inherently subjective.

Given this inherent ambiguity, we adopt a pragmatic, multi-perspective evaluation framework consisting of three complementary components:

\begin{itemize}
    \item \textbf{Evaluating alignment with TA scores.} We compare AI-assigned scores with those assigned by teaching assistants (TAs). While TA grading can be inconsistent and often lacks detailed personalized feedback, it is the practical reference in university assessments,  so alignment with TA practices is a necessary condition for deployment. Because the TA Gradescope rubric is designed for efficient human scoring, our AI rubric is adapted for automated interpretation and feedback; thus, we interpret score comparisons as measuring outcome alignment rather than strict rubric equivalence.

   \item \textbf{Direct student feedback.}
We return AI-generated scores and feedback to students (see Appendix~\ref{app:messagetostudent} for an example message) and collect survey responses from math 2A students on perceived correctness and usefulness. This helps identify potential misgrades and characterize how students interpret AI feedback, while recognizing that students may perceive the AI as overly harsh when its scores are lower than the TA-assigned scores.

    \item \textbf{Independent review by external evaluators.}
A team of trained graduate students, undergraduate students, and advanced high school students independently evaluates the AI outputs using the same rubrics applied by the AI, assessing both correctness and fairness and assigning their own scores. This provides an additional, partially decoupled, third-party  perspective beyond TA alignment and student feedback.

\end{itemize}
Due to practical, technical, and administrative constraints (e.g., segmentation failures, scan quality issues, unmatched test codes, and reviewer availability), not all quizzes could be included in every comparison; these exclusions were determined prior to evaluation and are independent of grading outcomes (i.e., non-selective). See Appendix~Section\ref{app:datacoverage} for a categorical  breakdown.

Also, we evaluate agreement across all free-response submissions without stratifying by solution correctness, since OCR reliability and rubric-based reasoning must function even when the underlying mathematics is correct; grading alignment is therefore nontrivial in both correct and incorrect cases.

\subsection{Alignment with TA Scores}
We evaluate agreement between an AI grader and human TAs on 3{,}945 handwritten free-response student-question records from the Math 2A and B. All Math 2B items are free response; in Math 2A, non-free-response items (e.g., multiple-choice or fill-in blanks) are excluded from this analysis. Each record consists of an AI score, a TA score, and the gap \(g = \mathrm{AI}-\mathrm{TA}\).

We summarize at two levels:
\begin{enumerate}
  \item Global distribution of the score gap across all records;
  \item Quiz-level aggregates: mean gap and percentage within 1 point (``within-1'');
\end{enumerate}

\subsubsection{Overall agreement (global)}
The overall gap distribution is sharply peaked at 0, meaning AI and TAs give the same or very similar scores for many submissions (Figure~\ref{fig:global-distribution}). The histogram also has small tails: most differences are near 0, but a few submissions have large gaps. On average, AI scores are 0.40 points lower than TA scores, with a standard deviation of 1.12 (about one point).
\begin{center}
  \includegraphics[width=0.45\linewidth]{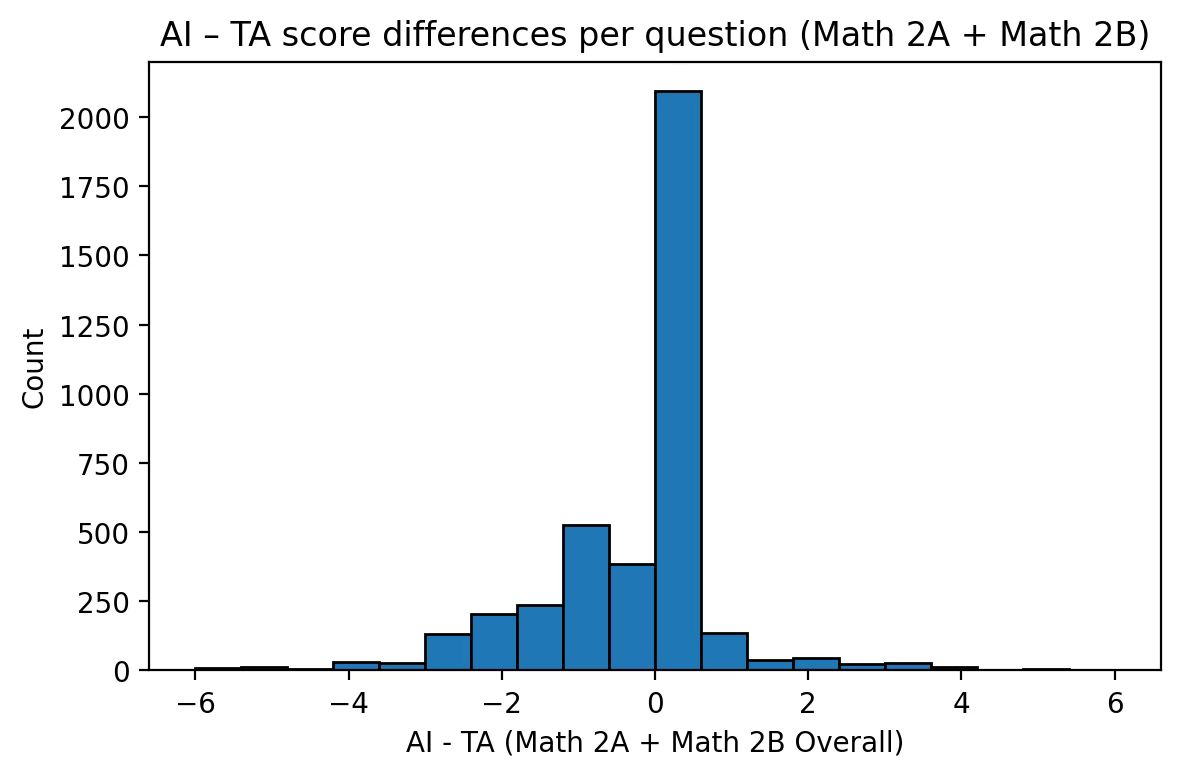}
  \captionof{figure}{Global distribution of AI--TA score gaps.}
  \label{fig:global-distribution}
\end{center}

\subsubsection{Quiz-level summary}
Across quizzes, the mean absolute gap (MAE) typically falls in the range \(0.5\) and \(1.06\) points, and the within-1 rate ranges from \(68\%\) to \(86\%\), demonstrating strong overall alignment with TA grading.

\begin{figure}[htbp]
  \centering
  \includegraphics[width=0.7\linewidth]{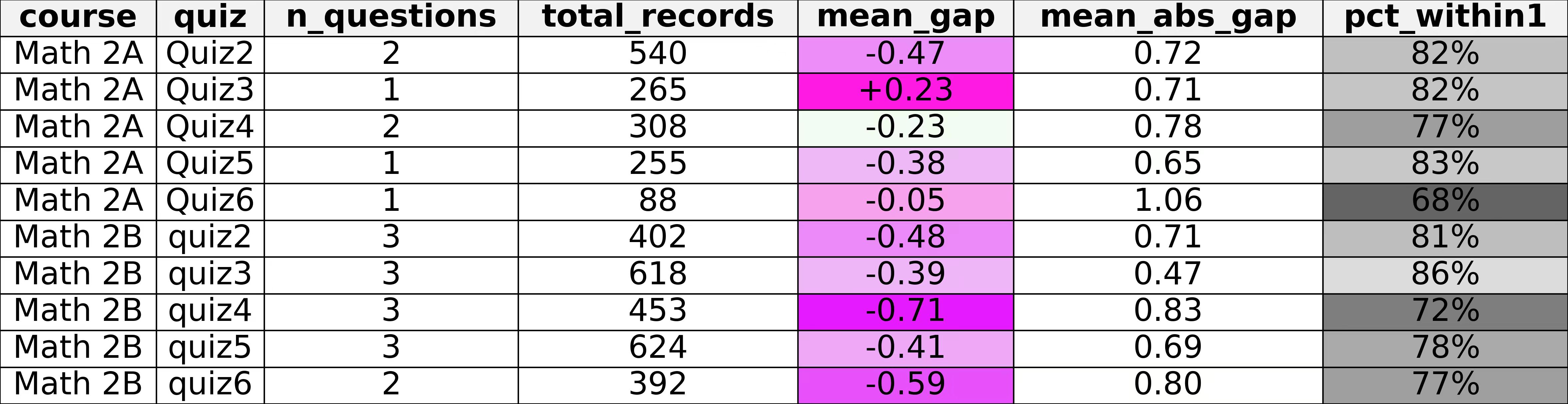}
  \caption{Quiz-level summary: mean gap and within-1 percentage.}
  \label{fig:quiz-level}
\end{figure}

For larger discrepancies with $|g|\ge 1$, negative gaps ($g\le -1$) are primarily attributable to OCR errors or occasional leniency in TA grading (i.e., the AI is comparatively harsher). For a representative positive outlier with $g>1$, see Appendix~\ref{app:ai-vs-ta-case}.

\subsection{Survey Result From Students}
The following summarizes results from a voluntary student survey regarding the feedback provided by our grading system. The survey was conducted in two sections of Math 2A (n = 301 respondents out of approximately 400 enrolled students). No survey was administered in Math 2B. The two independent sections exhibit highly consistent response patterns across all items.

\begin{center}
\setlength{\tabcolsep}{1pt}
\begin{tabular}{lcccc}
\hline
Course & Accurate\hspace{5pt}  & Clear\hspace{5pt}  & Prefer TA\hspace{5pt} & Open to AI  \\
\hline
2A Section 1 (n=158) & 60\% & 63\%  & 52\%  & 78\%  \\[1.5mm]
2A Section 2 (n=143) & 61\%  & 62\%  & 53\%  & 76\%  \\
\hline
\end{tabular}
\captionof{table}{Student survey summary. ``Accurate'' = feedback very/mostly accurate; ``Clear'' =feedback  very/mostly clear; ``Open to AI'' = yes or maybe with improvements.}
\end{center}

Overall, the survey suggests cautious openness toward AI-assisted grading, with many students reporting reasonable accuracy and clarity. Open-ended responses most often cited handwriting recognition (OCR), partial-credit strictness, and limited explanations, and emphasized the need for human oversight. These comments motivate robust OCR and an automatic review-triggering mechanism that escalates low-confidence cases for manual checking (Section~\ref{sec:limitation}).

\subsection{Agreement Between AI and Independent Human Reviewer}

While AI--TA score agreement was already high, score matching alone does not guarantee correctness or feedback quality. 
We therefore employed independent reviewers to review the AI outputs on 3{,}851 handwritten free-response student--question records (Math 2A: 1{,}819; Math 2B: 2{,}032) by verifying (i) input quality (OCR legibility), (ii) score alignment under the same grading rubric, and (iii) rubric fidelity and feedback accuracy. This provides an independent, rubric-consistent evaluation beyond raw TA alignment, which is valuable given the lack of an objective ground-truth score for handwritten free-response work.

\paragraph{Input Quality (OCR).}
Most scripts were readable: 87.64\% were rated \emph{Acceptable} and 12.36\% \emph{Problematic} (Figure~\ref{fig:ocr verdict}).
 see the OCR Section~\ref{sec:ocr-eval} for typical failure modes.

\paragraph{Score Agreement.}
Across 3851 graded responses, the score difference (AI–Reviewer) is sharply centered at~0 with light, symmetric tails; over 90\% lie within $\pm1$ point (Figure~\ref{fig:Score difference}), indicating strong numerical concordance.

\paragraph{Rubric Fidelity and Feedback Accuracy.}
Independent review found 79.79\% of AI evaluations fully correct, 9.55\% acceptable, and 10.67\% incorrect (Figure~\ref{fig:Grading overall verdict}).  Beyond OCR-related failures, two main error patterns emerged: (i) over-penalizing valid solution methods that are mathematically correct but not explicitly listed in the rubric, and (ii) undue leniency toward rubric-shaped explanations that follow familiar steps but contain incorrect reasoning.

\begin{center}
    \includegraphics[width=0.7\linewidth]{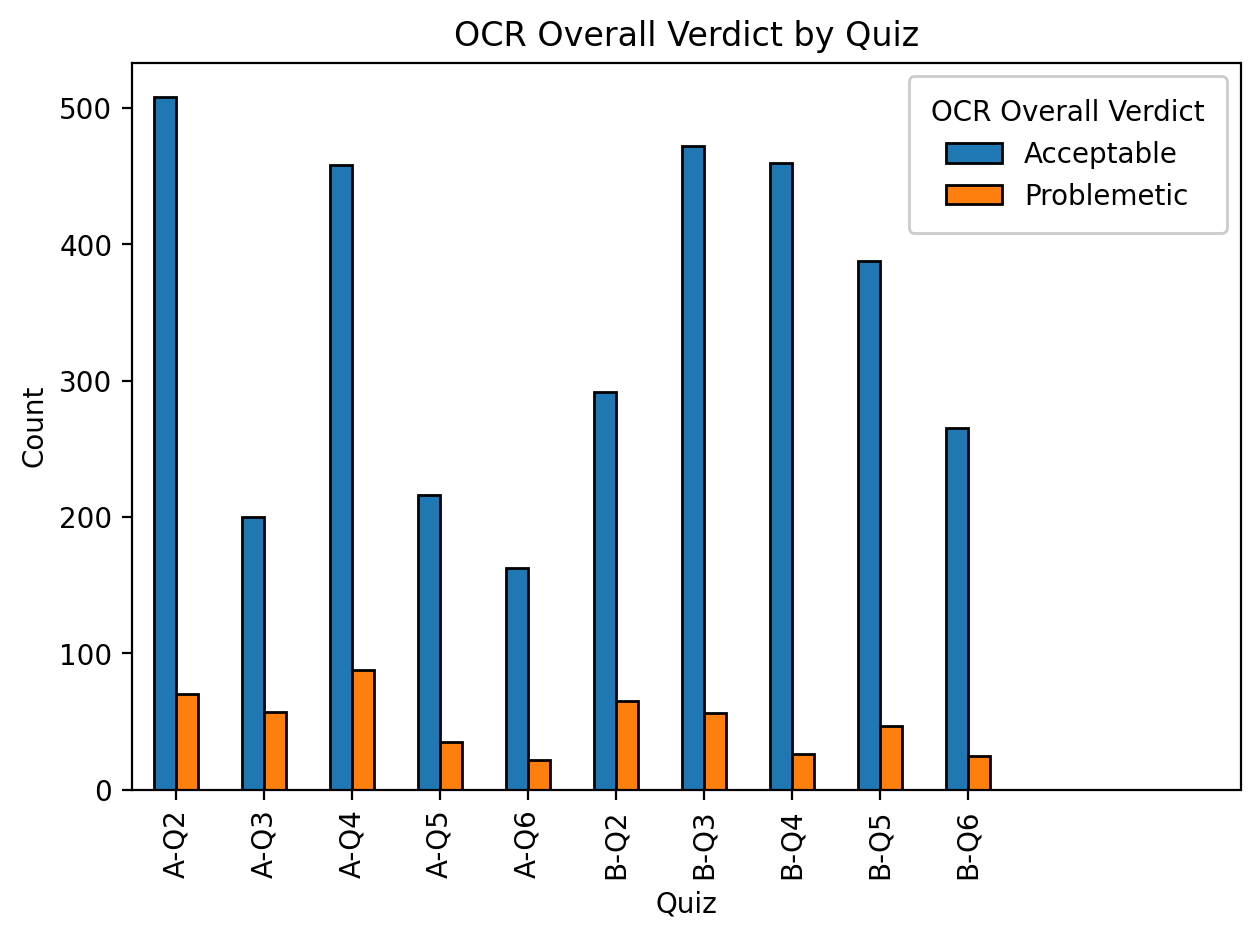}
    \captionof{figure}{OCR verdict distribution: A-Math 2A, B-Math 2B}
    \label{fig:ocr verdict}
 \end{center}

 \medskip
\begin{center}
    \includegraphics[width=0.7\linewidth]{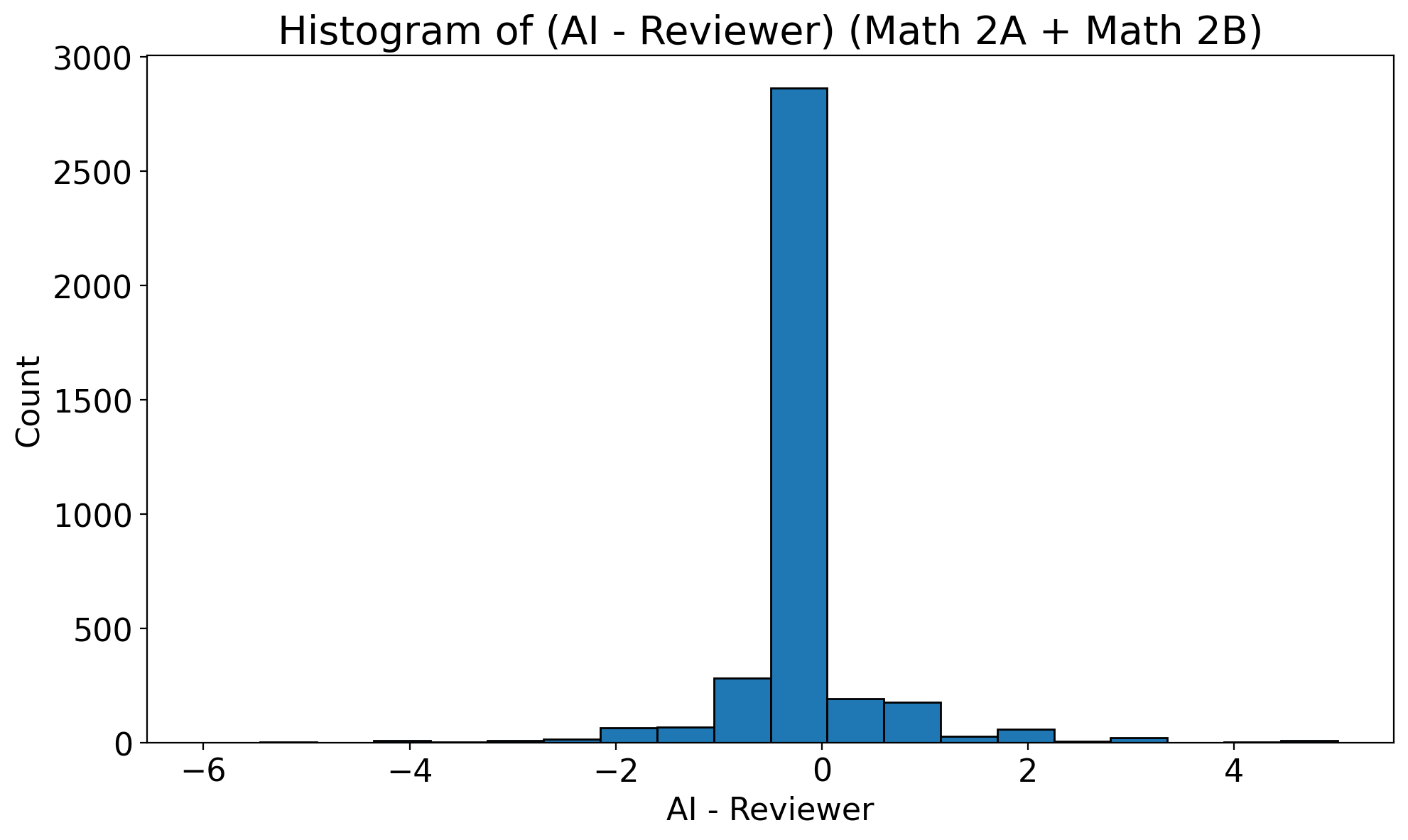}
    \captionof{figure}{AI–Reviewer score difference}
    \label{fig:Score difference}
\end{center}
  \begin{center}
    \includegraphics[width=0.8\linewidth]{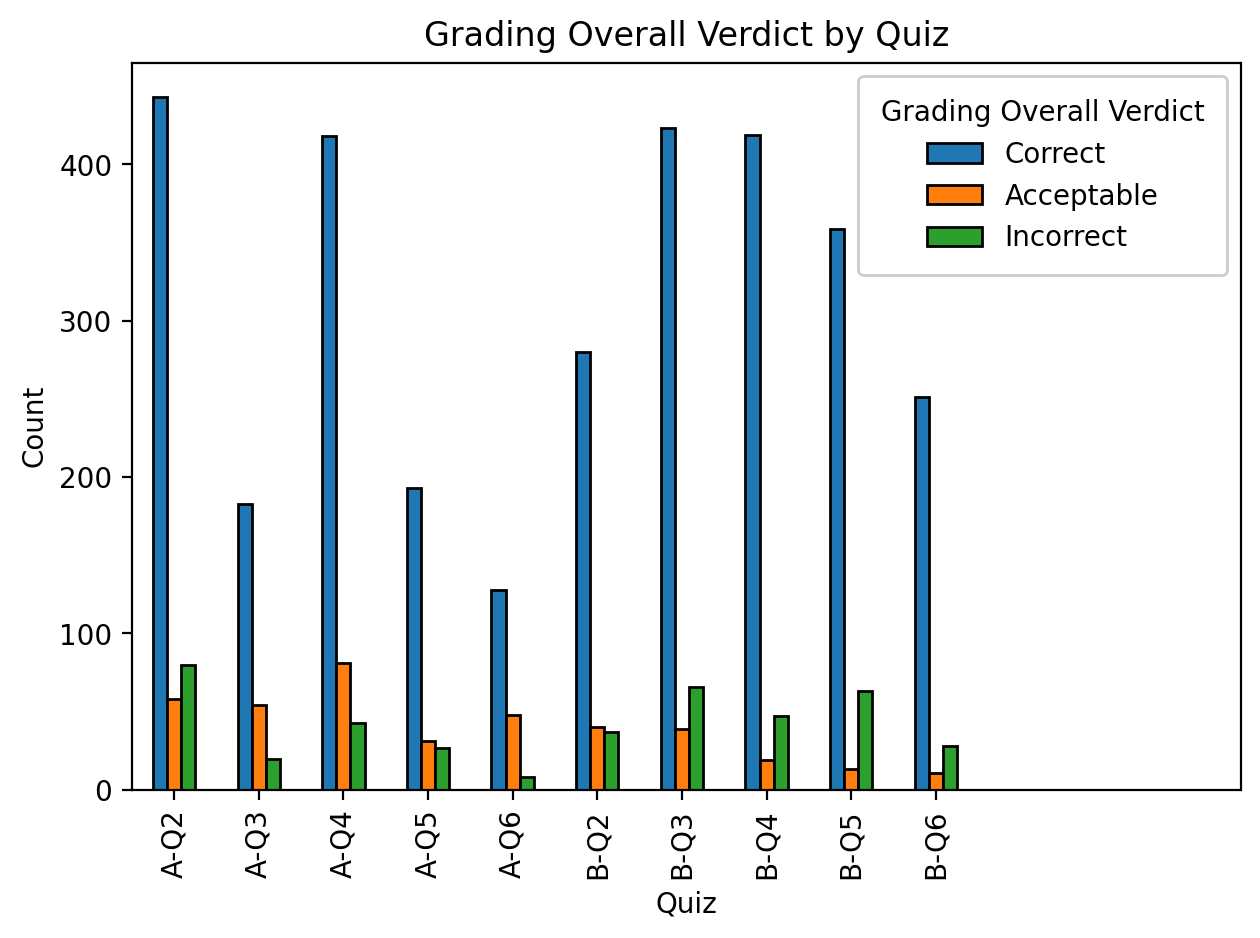}
    \captionof{figure}{Rubric fidelity and feedback audit:  A-Math 2A, B-Math 2B}
    \label{fig:Grading overall verdict}
  \end{center}

\section{Limitations}\label{sec:limitation}

Our study has several limitations that point to clear next steps.

\paragraph{Intervention policy.}
A key bottleneck for large-scale deployment is deciding when to trust automated grading versus defer to additional checks or human review. While OCR errors and rubric--prompt interactions remain major sources of failure, we do not yet have a systematic method for deciding when to trigger such checks or defer to human review. In our experiments, we flag only a narrow set of issues (e.g., a correct final answer receiving less than full credit, or high variance across repeated runs), and these triggers are simple, manually designed heuristics rather than a unified decision rule. Related ML work on learned deferral, uncertainty-aware risk-averse deployment, and fair abstention offers principled perspectives for triggering human review (e.g., \cite{madras2018,kiyani2025,casacuberta2025}), but adapting these frameworks to OCR-conditioned, partial-credit grading without ground truth remains open.

\paragraph{High-stakes deployment.}
Our experiments focus on low-stakes quizzes ($\approx$10\% of the course grade); extending AI-assisted grading to midterms and finals would raise the cost of errors and require substantially more conservative human-in-the-loop controls.

\paragraph{Choosing the most accurate grade/feedback.}
When multiple OCR engines or graders are available (e.g., GPT-based OCR vs.\ Mathpix, or different LLM graders), we currently rely on human reviewers to resolve disagreements and decide which output is more accurate. This does not scale and also slows our ability to evaluate or swap in new models. Developing a reliable automatic mechanism to detect disagreements and select the most reliable score$+$feedback is a future project.

\paragraph{Rubric aggregation (max-rule).}
We use the max-rule to aggregate rubric-level evidence, and it aligns best with TA grades in our experiments. However, we have not exhaustively compared it against other reasonable choices (e.g., averaging or other monotone score transforms), so conclusions about the optimal aggregation rule are specific to our setting.

\paragraph{Scope.}
Our evaluation focuses on single-variable calculus at one institution; additional deployments across courses and institutions are needed to establish generalizability and to understand operational impact under different grading cultures and student populations.

\section{Conclusion}
We presented a large-scale evaluation of AI-assisted grading for single-variable calculus at UC Irvine, covering thousands of quiz solutions from nearly 800 students. Our system integrates OCR, prompt-guided reasoning, and rubric-based scoring in an automated workflow, and was evaluated through comparisons with TA grades, structured human review, and student feedback.

Our results suggest that current AI systems can already serve as reliable grading assistants: they handle most routine problems accurately, provide usable feedback, and substantially reduce manual grading effort. Key challenges remain, including OCR failures on diagram- or geometry-heavy work and cancellations, consistent partial-credit assignment across diverse solution variants, and principled policies for triggering human review in low-confidence cases.

Looking ahead, we plan to release well-documented benchmark subsets from our collected database to support reproducible evaluation of OCR and rubric-based grading. Specifically, we will provide two tracks: {\bf Track A} (clean) with corrected transcriptions and instructor/TA–approved reference grades, and {\bf Track B} (noisy) retaining original OCR outputs and reviewer labels to assess robustness and inform triage (i.e., when to defer to human review). We hope these resources will enable more transparent, scalable comparisons and accelerate progress toward reliable automated assessment in large-enrollment STEM courses.

\section*{Acknowledgments}
We gratefully acknowledge the UC Irvine School of Physical Sciences and the Department of Mathematics for their continued support of this AI grading project. We are deeply grateful to our graduate student reviewers—Jongwon D. Kim, Yuanduan Lou, Shiyi Luo, and Zeyi Xu; our undergraduate student reviewers—Keith Bui, Xin Gao, Changze Li, Tony Li, Alex Lu, Eric Lyu, Xuaner Niu, Oleg Presnyakov, Scott Sibley, Vincent Wang, Qian Zhang, and Mickey Zhao; and our high school student reviewers—Eric Guo, Kevin Huang, Aaron Liu, Jasper Sun, An My Ngoc Tran, and Lauren Zhang—for reviewing AI grading results and providing constructive feedback and suggestions. We especially thank Oleg Presnyakov for suggesting the max-rule (selecting the maximum score across multiple rubrics). Special thanks are due to the MATH 2A/2B instructors—Rachel Cohen Lehman, Ziming Shi, Daniel V. Simonson, and Yang Zhang—for their generous support, and to the MATH 2A/2B teaching assistants for their valuable assistance throughout this project. Long Chen was partially supported by NSF grant DMS-2309777, Jack Xin was partially supported by NSF grant DMS-2309520, and Yifeng Yu was partially supported by NSF grant DMS-2000191. We used LLM-based language tools to assist with wording and presentation (including rephrasing and structural edits); the authors are solely responsible for all technical content and conclusions.
\bibliographystyle{unsrt}
\bibliography{reference_paper}

\newpage
\appendix

\section{Appendix}
\subsection{Dataset Coverage Across Validation Protocol}\label{app:datacoverage}
Due to practical technical and administrative constraints, not all submitted quizzes could be included in every validation protocol. We distinguish (i) \emph{universal} exclusions that apply to any downstream analysis, and (ii) additional constraints specific to particular protocols (e.g., AI--TA score comparison or independent human review). Exclusions were driven by processing or logistical feasibility rather than by grading outcomes, and are intended to be non-selective with respect to score disagreement.

\paragraph{Universal exclusions (apply to all protocols).}
First, some quizzes were not successfully processed by our pipeline (most commonly due to segmentation failures when students wrote outside designated boxes, scan-quality issues, or occasional OCR/input-extraction failures). In addition, as part of model iteration and coordination with course staff, the first quiz in both Math 2A and Math 2B was excluded from our evaluation sets. Finally, one Math 2B TA section exhibited systematic scanning artifacts (e.g., extraneous lines on scanned pages) that materially degraded segmentation and input extraction; we therefore excluded that section from all analyses.

\paragraph{Additional constraints for AI--TA score comparison (record linkage).}
The AI--TA comparison further requires linking each processed submission to a Gradescope record via the assigned de-identification test code. This linkage sometimes fails because a subset of students omitted or mistyped the code, and in some cases TA score exports from Gradescope were incomplete or missing for reasons outside our control, preventing a reliable per-question score comparison.

\paragraph{Additional constraints for independent human review (audit logistics).}
The independent review was organized at the TA-section level: each reviewer reviewed quizzes from a single TA section, and not all TA sections could be covered; for example, when a TA taught multiple discussion sections, we sometimes reviewed only a subset of those sections due to limited reviewer capacity. A small fraction of assigned audits were not completed because some reviewers did not finish them.

\subsection{Design of Answer Sheets}\label{section:answersheet}
Below are two answer-sheet designs. Version 1 allows blank answers, which can trigger hallucinations despite explicit OCR instructions. Version 2 prevents this by placing ``{\bf Solution}” and ``{\bf Final Answer}” in designated boxes.

\begin{center}
  \includegraphics[width=0.26\linewidth]{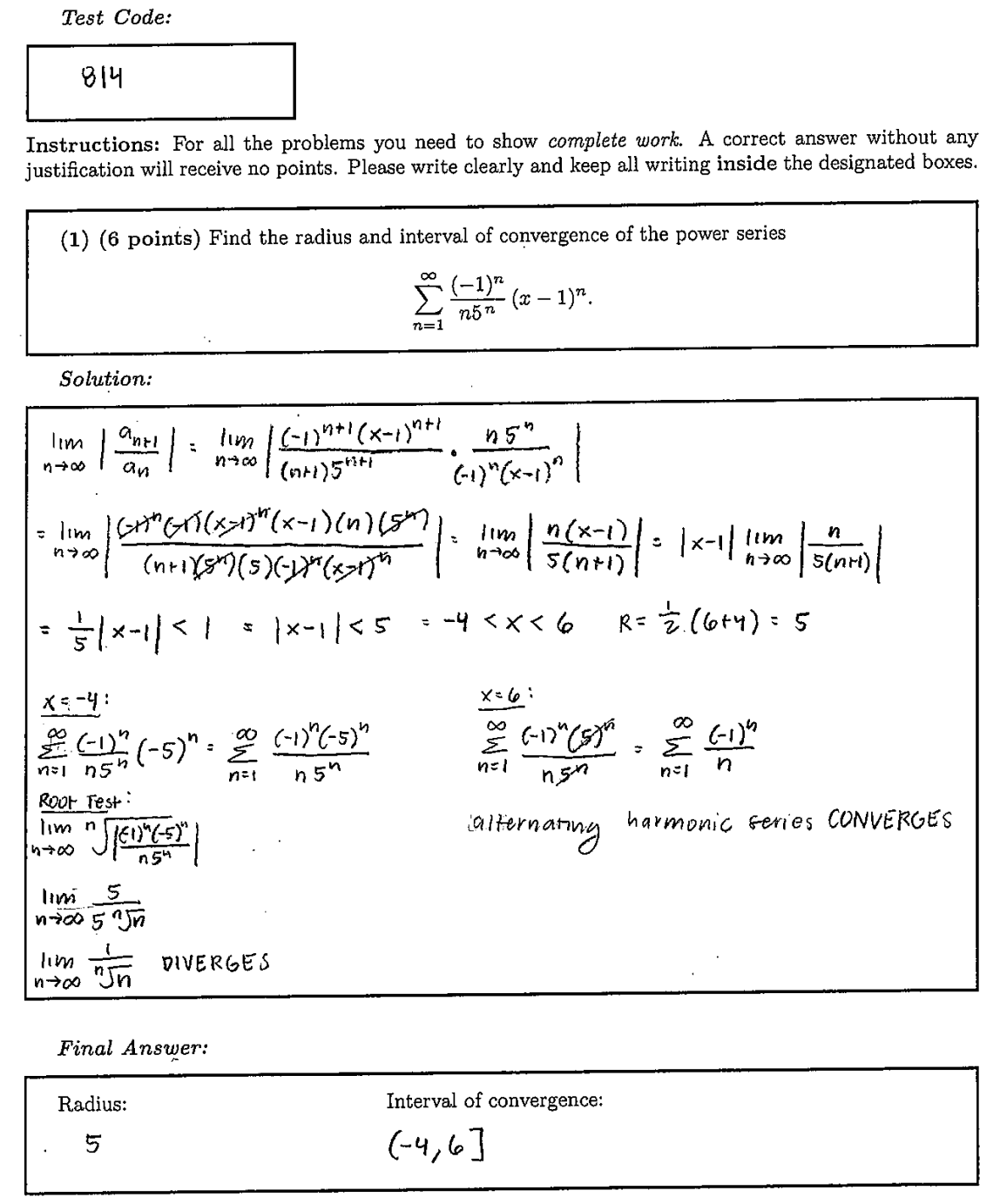}
\qquad
  \includegraphics[width=0.26\linewidth]{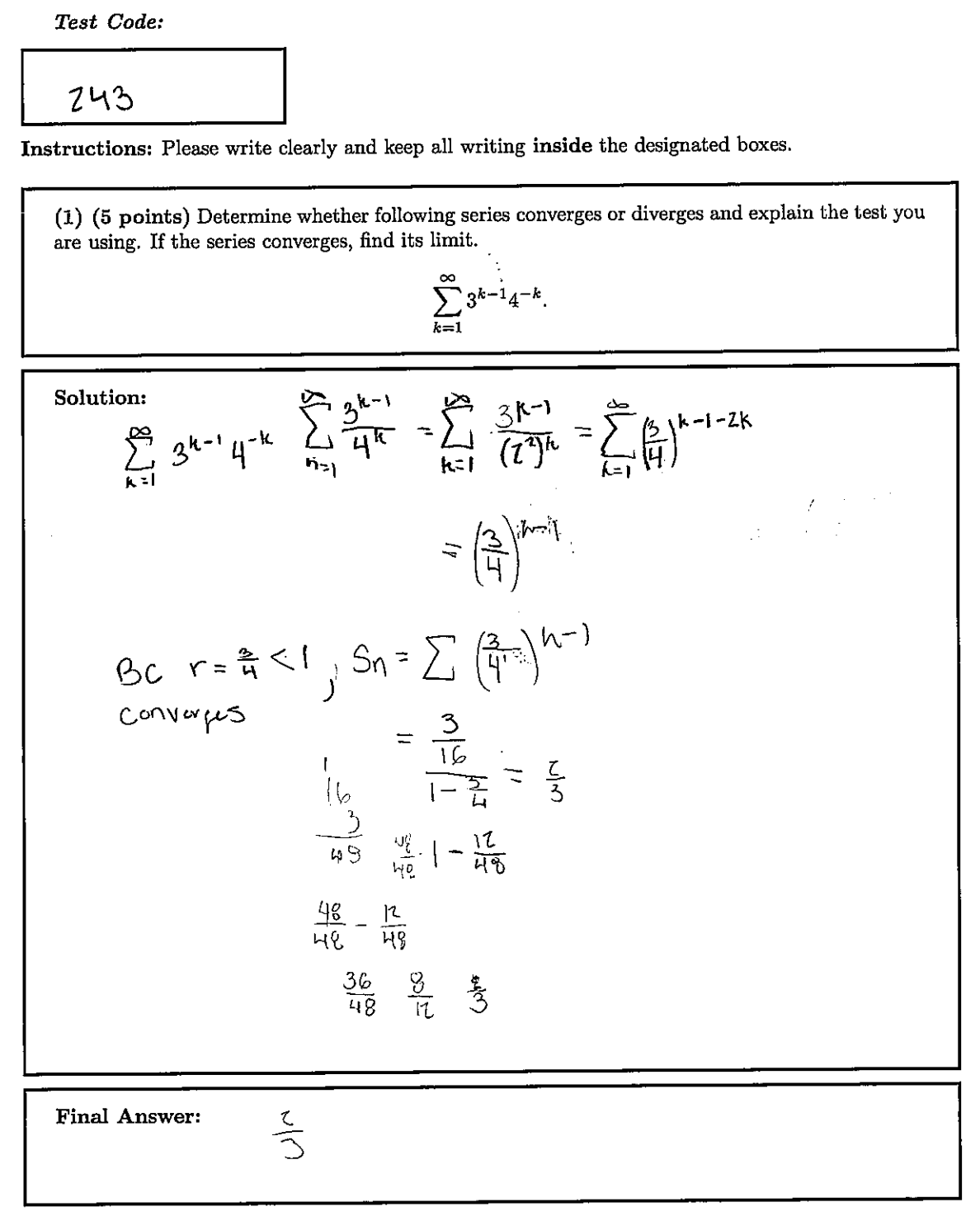}
  \captionof{figure}{Left: Answer Sheet Version 1, Right: Answer Sheet Version 2.}
\end{center}

\subsection{GPT-4.1-mini vs. Mathpix}\label{section:gptvsmathpix}
In what follows, we present several representative examples showing that (1) ChatGPT~4.1-mini demonstrates strong performance, (2) illustrative comparisons between Mathpix and ChatGPT 4.1 mini outputs  highlight key differences, and (3) how to effectively mitigate issues such as autocorrection and hallucination.

\begin{example}\label{example:r-sum}
\emph{Read the following handwritten solution from a Math 2B quiz. The problem is to find the sum of the series:}
\[
\sum_{n=3}^{\infty} \frac{2^{\,n-1}}{3^{\,n+1}} .
\]
\begin{center}
 \includegraphics[scale=0.35]{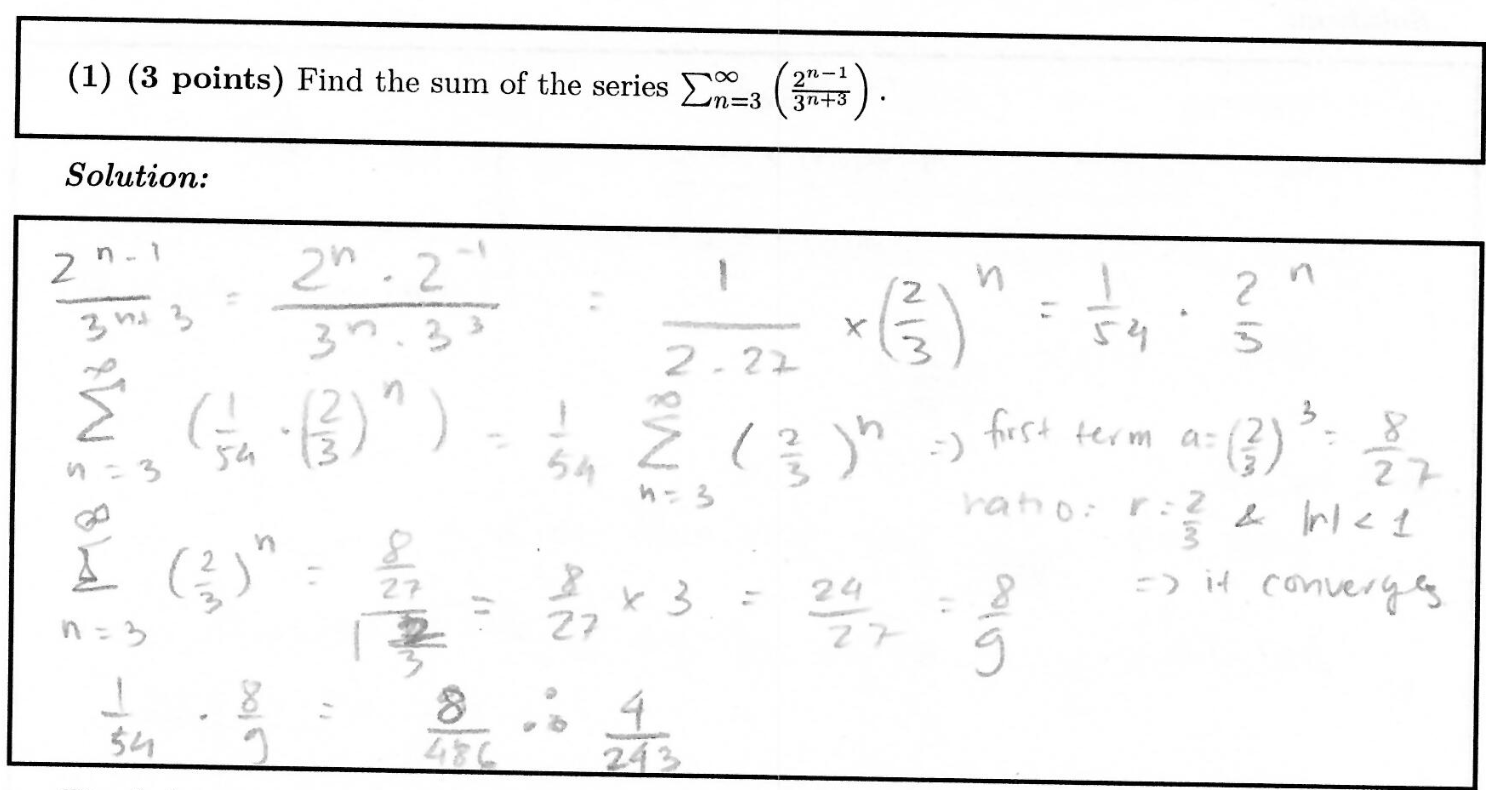}
\end{center}
\emph{\textbf{OCR output from GPT-4.1-mini, returned as LaTeX via our API pipeline and rendered here as a compiled PDF.} The transcription is nearly perfect and the correct score is assigned by the downstream evaluation.}
\[
\frac{2^{n-1}}{3^{n+3}} = \frac{2^n \cdot 2^{-1}}{3^n \cdot 3^3} = \frac{1}{2 \cdot 27} \times \left(\frac{2}{3}\right)^n = \frac{1}{54} \cdot \left(\frac{2}{3}\right)^n
\]
\[
\sum_{n=3}^\infty \left(\frac{1}{54} \cdot \left(\frac{2}{3}\right)^n \right) = \frac{1}{54} \sum_{n=3}^\infty \left(\frac{2}{3}\right)^n \quad \Rightarrow \text{first term } a = \left(\frac{2}{3}\right)^3 = \frac{8}{27}
\]
ratio: $r = \frac{2}{3}$ \& $|r| < 1$
\[
\sum_{n=3}^\infty \left(\frac{2}{3}\right)^n = \frac{\frac{8}{27}}{1 - \frac{2}{3}} = \frac{8}{27} \times 3 = \frac{24}{27} = \frac{8}{9} \quad \Rightarrow \text{it converges}
\]
\[
\frac{1}{54} \cdot \frac{8}{9} = \frac{8}{486} = \frac{4}{243}
\]

\emph{This example demonstrates the reasoning-based reading capability of 4.1-mini. From the surrounding context, the model correctly inferred the student’s intended expressions: (1) it repaired a minor notational omission by inserting missing parentheses, interpreting the term as $\left(\frac{2}{3}\right)^n$; and (2) it resolved an ambiguous notation in the denominator (the minus sign is faint or possibly absent) of $\frac{\frac{8}{27}}{1-\frac{2}{3}}$ rather than interpreting it as the mixed number $1\frac{2}{3}$. We therefore classify the output as \emph{almost perfect}, reflecting correct intent recovery rather than literal transcription. In this case, a strictly literal reading could lead to unwarranted point loss during grading.}

\medskip

{\bf OCR output from Mathpix, returned as LaTeX via our API pipeline and rendered here as a compiled PDF.:} The transcription contains  several typos, which leads to unwanted one point loss in the downstream evaluation part.  

\medskip

$\begin{array}{l}\frac{2^{n-1}}{3^{n+3}}=\frac{2^{n} \cdot 2^{-1}}{3^{n} \cdot 3^{3}}=\frac{1}{2 \cdot 27} \times\left(\frac{2}{3}\right)^{n}=\frac{1}{54} \cdot \frac{2}{3} n \\[3mm] \sum_{n=3}^{\infty}\left(\frac{1}{54} \cdot\left(\frac{2}{3}\right)^{n}\right)=\frac{1}{54} \sum_{n=3}^{\infty}\left(\frac{2}{3}\right)^{n} \Rightarrow \text { firsterm } a=\left(\frac{2}{3}\right)^{3}=\frac{8}{27} \\[3mm] \frac{k}{\infty}\left(\frac{2}{3}\right)^{n}=\frac{8}{27}=\frac{8}{27} \times 3=\frac{24}{27}=\frac{8}{9} \quad \Rightarrow \text { it converges } \\[3mm] n=3 \quad \frac{8}{34} \cdot \frac{8}{9}=\frac{8}{486} \cdot \frac{4}{243} \\[3mm] \frac{1}{54}\end{array}$

\medskip

\emph{Out of curiosity, we also processed the above sample on Mathpix Snip, the online version. Below is one output (rendered in PDF). }

\medskip

$\begin{aligned} & \frac{2^{n-1}}{3^{n+3}}=\frac{2^n \cdot 2^{-1}}{3^n \cdot 3^3}=\frac{1}{2 \cdot 27} \times\left(\frac{2}{3}\right)^n=\frac{1}{54} \cdot \frac{2}{3} \\ & \sum_{n=3}^{\infty}\left(\frac{1}{54} \cdot\left(\frac{2}{3}\right)^n\right)=\frac{1}{54} \sum_{n=3}^{\infty}\left(\frac{2}{3}\right)^n \Rightarrow \text { first term } a=\left(\frac{2}{3}\right)^3=\frac{8}{27} \\ & \frac{k}{n}\left(\frac{2}{3}\right)^n=\frac{\frac{8}{27}}{1 \frac{2}{3}}=\frac{8}{27} \times 3=\frac{24}{27}=\frac{8}{9} \quad \Rightarrow \text { it converges } \\ & \frac{1}{54} \cdot \frac{8}{9}=\frac{8}{486} \therefore \frac{4}{243}\end{aligned}$
\end{example}

\bigskip

\begin{example}
\emph{Below is another example demonstrating the strength of 4.1-mini in reading messy handwritten English words. The problem is: 
\noindent
\textbf{(4 points)} Determine whether the series
\[
\sum_{n=1}^{\infty} \frac{(-1)^n}{\sqrt{n}}
\]
converges or diverges. Show all work and state the test you are using.}

\begin{center}
 \includegraphics[scale=0.35]{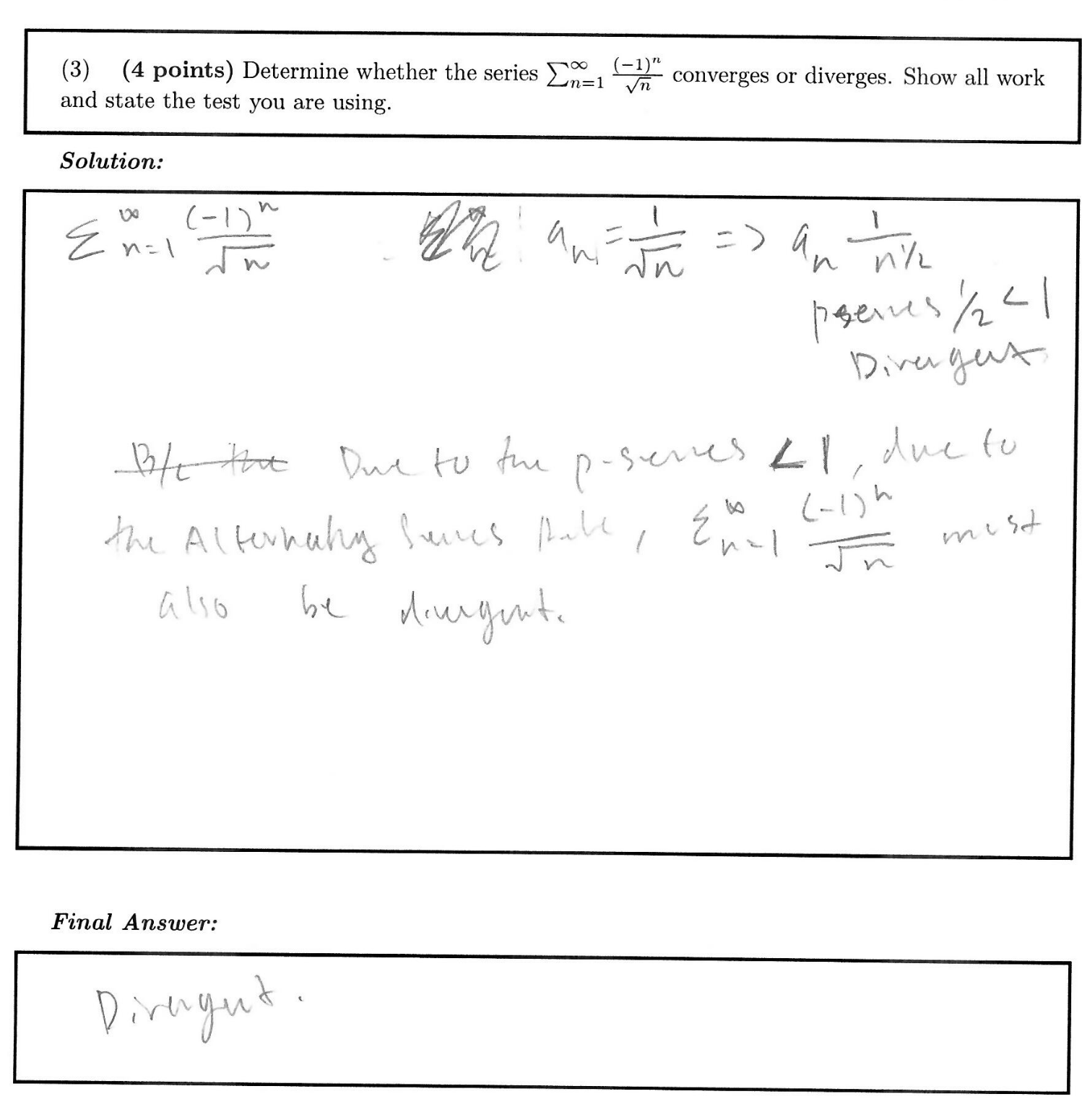}
\end{center}

\emph{Note that the handwritten English words are not easy to recognize, even to human readers.}

\medskip

{\bf OCR output from GPT-4.1-mini, returned as LaTeX via our API pipeline and rendered here as a compiled PDF:} The transcription is almost perfect: it is perfect in the solution part, with a single typo in the final answer (likely due to the lack of context).

\medskip
\noindent
\textbf{Solution Part:}
\[
\sum_{n=1}^{\infty} \frac{(-1)^n}{\sqrt{n}} \quad a_n = \frac{1}{\sqrt{n}} \implies a_n \frac{1}{n^{1/2}}
\]
p-series \( \frac{1}{2} < 1 \) Divergent

Due to the p-series \(\leq 1\), due to the Alternating Series Rule, \(\sum_{n=1}^{\infty} \frac{(-1)^n}{\sqrt{n}}\) must also be divergent.

\medskip
\noindent
\textbf{Final answer:} \text{Dirugnt.}

\medskip
\noindent
\emph{Although the final answer contains a typo,  the downstream evaluation module still correctly interpret ``Dirugnt" as a misspelling of ``divergent".}

\medskip
\noindent
{\bf OCR output from Mathpix, returned as LaTeX via our API pipeline and rendered here as a compiled PDF:} \emph{In contrast, the Mathpix transcription contains many typos in both mathematical notation and English words:}

\medskip
\noindent
\textbf{Solution part:} $\sum_{n=1}^{\infty} \frac{(-1)^{n}}{\sqrt{n}} \quad$ an $a_{n}=\frac{1}{\sqrt{n}} \Rightarrow a_{n} \frac{1}{n^{1 / 2}}$ penns $1 / 2<1$ Divergent

B/c Dare to the p-serves $<1$, duc to the Alternahy sums thate, $\sum_{n=1}^{\infty} \frac{(-1)^{n}}{\sqrt{n}}$ mist also be dougunt.

\medskip
\noindent
\textbf{Final answer part:} Direngut.

\emph{We would like to remark that despite the numerous word-level typos, the downstream model is still able to recover the student’s intended reasoning from context. Nevertheless, there are examples where confusion caused by misspelling did contribute unwanted point loss. }
\end{example}

\medskip

\begin{example}
    \emph{This example shows that GPT-4.1-mini can correctly read very faint handwriting and produce a correct downstream interpretation, whereas Mathpix fails to do so, leading to unwarranted point loss in the downstream evaluation.}

\begin{center}
 \includegraphics[scale=0.3]{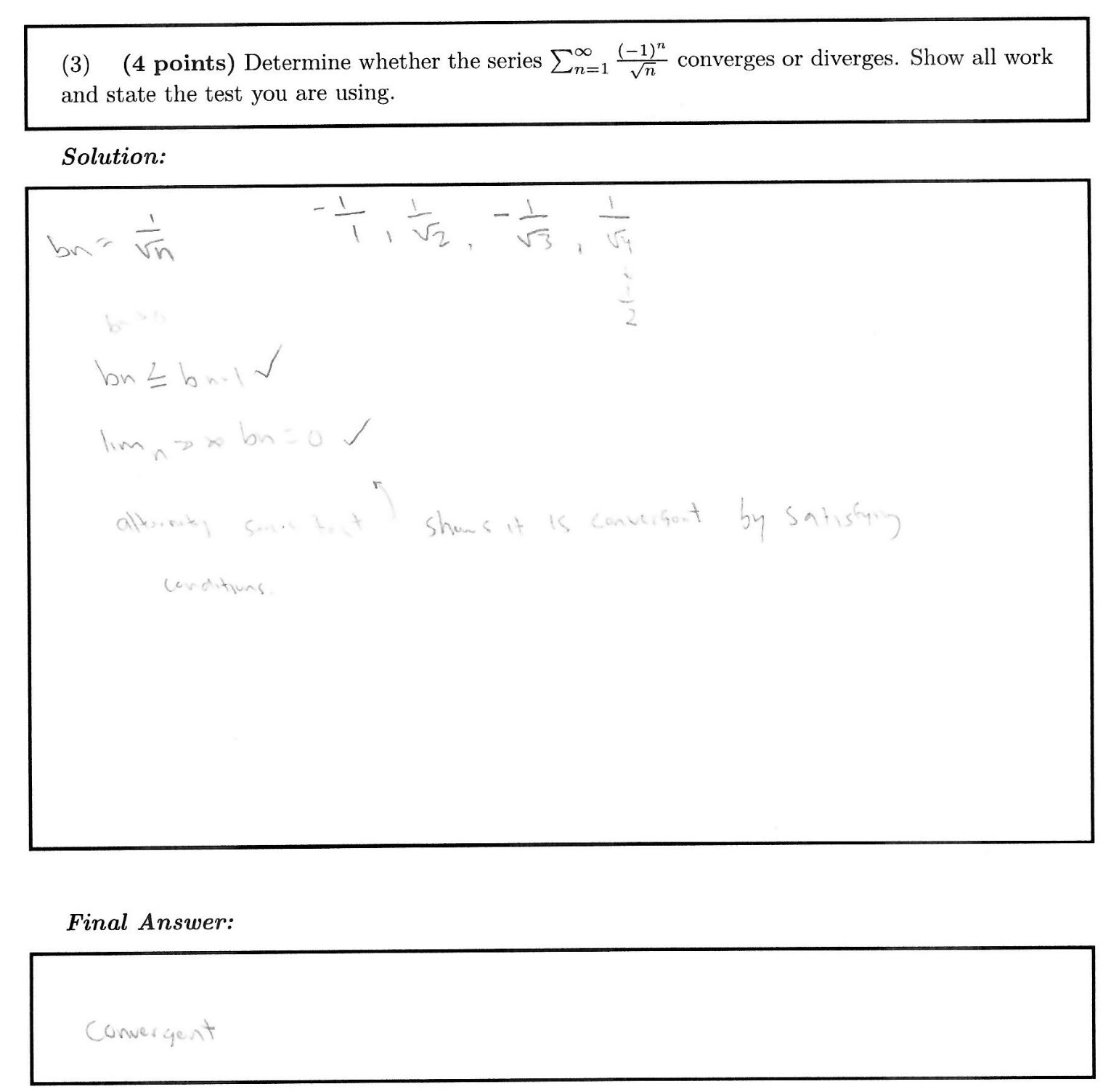}
\end{center}

{\bf OCR output from GPT-4.1-mini, returned as LaTeX via our API pipeline and rendered here as a compiled PDF.}: \emph{The transcription is perfect.}

\medskip
\noindent
{\bf Solution Part:}
\[
b_n = \frac{1}{\sqrt{n}} \quad -\frac{1}{1}, \frac{1}{\sqrt{2}}, -\frac{1}{\sqrt{3}}, \frac{1}{\sqrt{4}}
\]

\[
b_n \leq b_{n-1} \checkmark
\]

\[
\lim_{n \to \infty} b_n = 0 \checkmark
\]

alternating series test \(\Rightarrow\) shows it is convergent by satisfying conditions.

\medskip

\noindent
{\bf Final answer Part:} Convergent

\medskip

{\bf OCR output from Mathpix, returned as LaTeX via our API pipeline and rendered here as a compiled PDF}: \emph{The transcription contains multiple typos in both mathematical notation and English words, and it misses a key part of the test name, which results in a 1-point loss (out of 4) in the downstream evaluation.}

\medskip
\noindent
{\bf Solution Part:}
\[
\begin{array}{l}
b_{n}=\frac{1}{\sqrt{n}} \quad-\frac{1}{1}, \frac{1}{\sqrt{2}},-\frac{1}{\sqrt{3}}, \frac{1}{\sqrt{4}} \\
b_{n} \leq b_{n-1} \sqrt{ } \\
\lim _{n \rightarrow \infty} \rightarrow b_{n}=0 /
\end{array}
\]
\[
\begin{array}{l}
\text { curcitiums }
\end{array}
\]

\medskip

\noindent
{\bf Final Answer Part:} Convergent

\end{example}

\subsection{Failure Example of 4.1-mini OCR}\label{app:4.1-ocr-fail}  The problem is 
\noindent
\textbf{ (3 points)} Find the sum of the series
\[
\sum_{n=3}^{\infty} \frac{2^{\,n-1}}{3^{\,n+1}} .
\]

\begin{center}
 \includegraphics[scale=0.5]{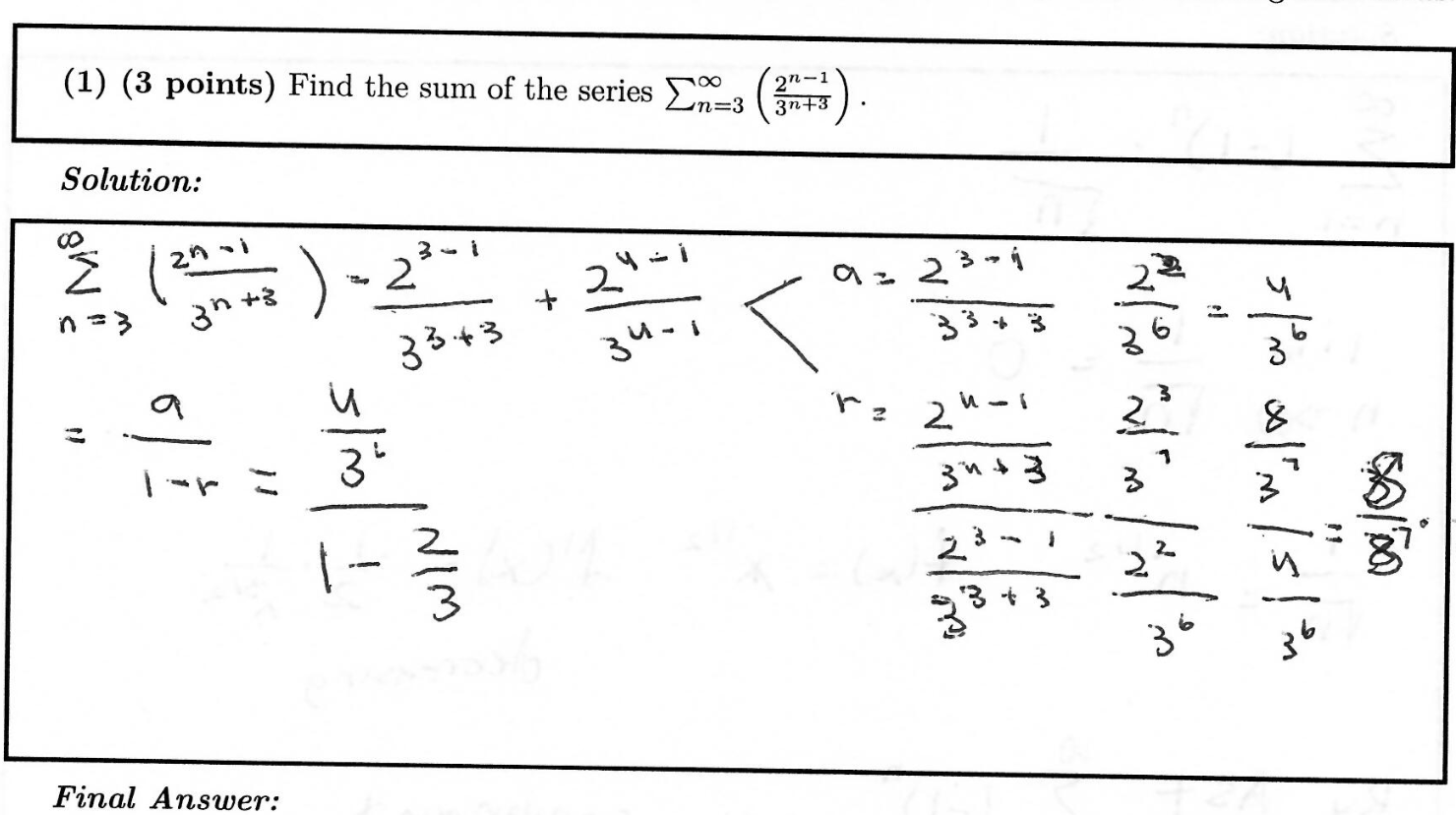}
 \captionof{figure}{Representative of the dominant division-related failure mode}
\end{center}

\textbf{OCR output from GPT-4.1-mini, returned as LaTeX via our API pipeline and rendered here as a compiled PDF:} This output below exhibits two issues. 
(1) \emph{Missing content:} part of the fraction expression is omitted, likely due to the presence of multiple nested fraction layers. 
(2) \emph{Unwanted autocorrection:} the model alters an exponent in a way that is not visually justified, changing what the student wrote,
$2^{4-1}/3^{4-1}$, into $2^{4-1}/3^{4+3}$, which is mathematically consistent in this context but does not faithfully reflect the student’s written work.

{\bf Solution Part:}
\[
\sum_{n=3}^{\infty} \left(\frac{2^{n-1}}{3^{n+3}}\right) = \frac{2^{3-1}}{3^{3+3}} + \frac{2^{4-1}}{3^{4+3}} \quad \leftarrow \quad a = \frac{2^{3-1}}{3^{3+3}} \quad \frac{2^2}{3^6} = \frac{4}{3^6}
\]

\[
r = \frac{2^{n-1}}{3^{n+3}} \quad \frac{2^3}{3^7} \quad \frac{8}{3^7} \quad \frac{8}{3^7}
\]

\[
= \frac{a}{1-r} = \frac{\frac{4}{3^6}}{1-\frac{2}{3}}
\]

\subsection{Autocorrection Issue}\label{sec:autocorrection}
In this example, we illustrate the importance of explicitly instructing the model ``{\it Do not correct any math or logical error or typos}" (OCR prompt, see Section~\ref{sec:ocr-eval}) during the OCR stage to reduce unwanted autocorrection.

\medskip

\begin{example}\label{example:autocorrection-limit}
\emph{Consider the following handwritten solution from a Math 2A quiz: Evaluate the limit, if it exists. (If an answer does not exist, write DNE.) You need to show complete work. A correct answer without any justification will receive no points.}
\[
\lim_{x \to 3^+} \frac{x^2 - x - 6}{|x - 3|}.
\]

\begin{center}
 \includegraphics[scale=0.5]{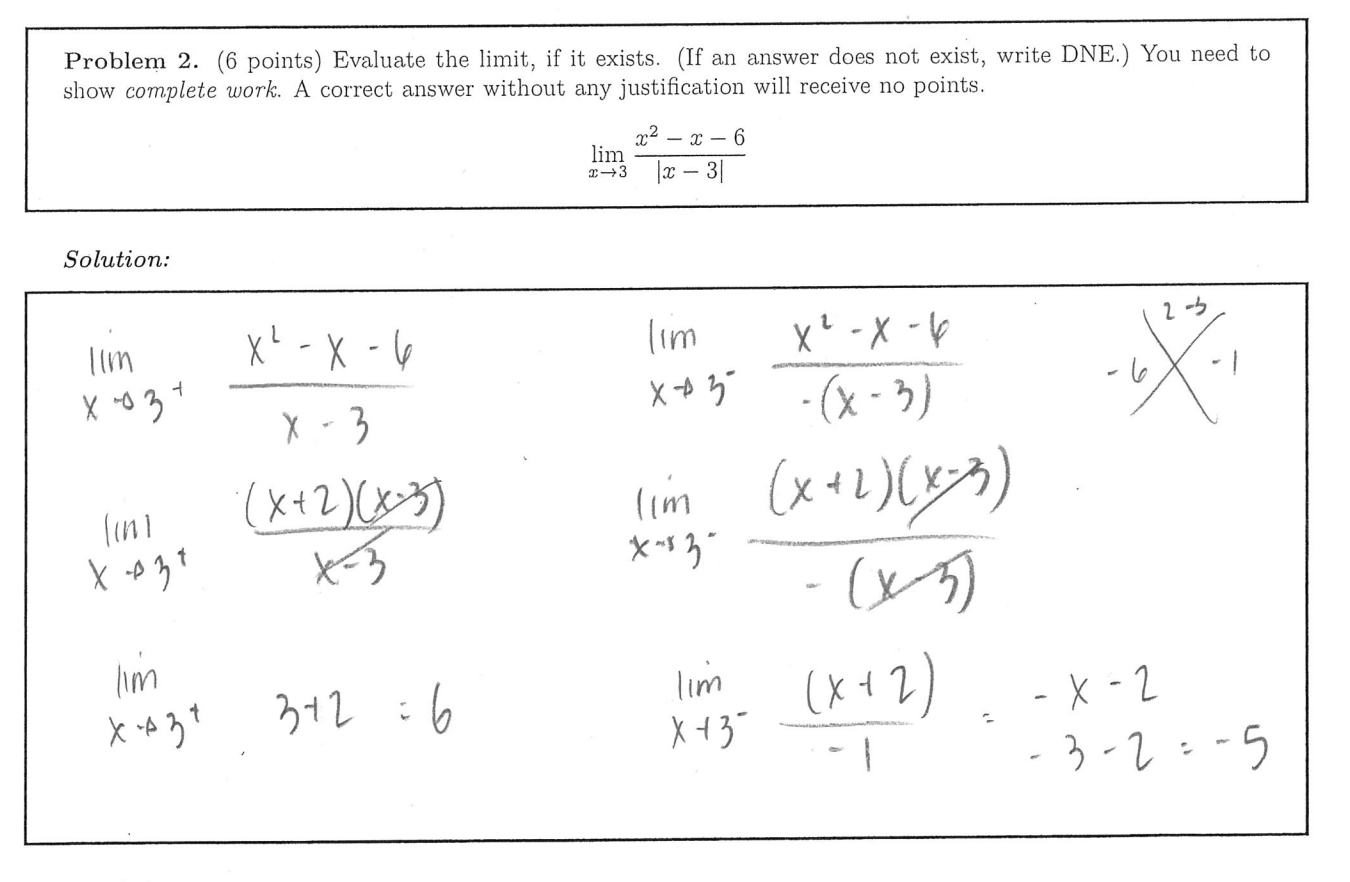}
\end{center}

\emph{Note that the student makes an obvious arithmetic mistake, writing $3+2=6$. We ran the GPT-4.1-mini OCR and transcription prompt three times {\bf without} including the  ``{\it Do not correct}” instruction:}

\begin{center}
\begin{minipage}[t]{0.32\textwidth}
\[
\begin{aligned}
{\text{\bf Output 1:}} \ \lim_{x\to3^+} \frac{x^2-x-6}{x-3} \\
= \lim_{x\to3^+} \frac{(x+2)(x-3)}{x-3} \\
= \lim_{x\to3^+} (x+2) = 3+2=6 \\[1ex]
\lim_{x\to3^+} \frac{x^2-x-6}{|x-3|} \\
= \lim_{x\to3^+} \frac{x^2-x-6}{-(x-3)} \\
= \lim_{x\to3^+} \frac{(x+2)(x-3)}{-(x-3)} \\[1ex]
\lim_{x\to3^+} \frac{x+2}{-1}
= \lim_{x\to3^+} -(x+2)\\ = -(3+2)=-5
\end{aligned}
\]
\end{minipage}
\hfill
\begin{minipage}[t]{0.32\textwidth}
\[
\begin{aligned}
{\text{\bf Output 2:}} \ \lim_{x\to3^+} \frac{x^2-x-6}{x-3} \\
= \lim_{x\to3^+} \frac{(x+2)(x-3)}{x-3} \\
\lim_{x\to3^+} (x+2) = 3+2=6 \\[1ex]
\lim_{x\to3^-} \frac{x^2-x-6}{|x-3|} \\
= \lim_{x\to3^-} \frac{(x+2)(x-3)}{-(x-3)} \\[1ex]
\lim_{x\to3^-} \frac{x+2}{-1}
= -(x+2) \\
= -3-2=-5
\end{aligned}
\]
\end{minipage}
\hfill
\begin{minipage}[t]{0.32\textwidth}
\[
 \begin{aligned}
{\text{\bf Output 3:}}\ \lim_{x\to3^+} \frac{x^2-x-6}{x-3} \\
= \lim_{x\to3^+} \frac{(x+2)(x-3)}{x-3} \\
= \lim_{x\to3^+} (x+2) \\
= {\bf 3+2=5} \\[1ex]
\lim_{x\to3^-} \frac{x^2-x-6}{-(x-3)} \\
= \lim_{x\to3^-} \frac{(x+2)(x-3)}{-(x-3)} \\
= \lim_{x\to3^-} \frac{x+2}{-1} \\
= -x-2 \\
= -3-2=-5
\end{aligned}
\]
\end{minipage}
\end{center}

\emph{As we can see, on the third run the AI silently corrected the student’s error:
\[
3+2=6 \quad \Rightarrow \quad {3+2=5}.
\]
This type of implicit autocorrection can become a serious issue when grading hundreds of tests, as it alters the original student work.}

\emph{After we added the instruction “Do not correct any math or logical errors or typos” to the OCR prompt, this autocorrection behavior no longer occurred, even after multiple runs.}

\emph{We note that unwanted autocorrection errors cannot be entirely eliminated (see the example shown in Section~\ref{app:4.1-ocr-fail}). In our experiments, such cases were rare, occurring in fewer than 2\% of samples.}

\end{example}

\subsection{Examples Illustrating the Roles of System Prompts}\label{app:systemmessage}

In this section, we present two examples to illustrate the roles of Prompts 1, 2, and 3 in the system message (Section~\ref{sec:systemmessage}).

The first example concerns Prompts 1 and 2, which emphasize logical consistency in the evaluation.

The first one is about Prompt 1 and 2 that stress the logic consistency in the system message. 

\begin{example} \emph{The following is a sample from Math 2B. The problem is to integrate}
\[
\int \cos^3 t \, \sin^2 t \, dt .
\]
\emph{A standard approach is to use the substitution}
\[
u=\sin t.
\]
\emph{Below is an incomplete student solution:}
\begin{center}
 \includegraphics[scale=0.4]{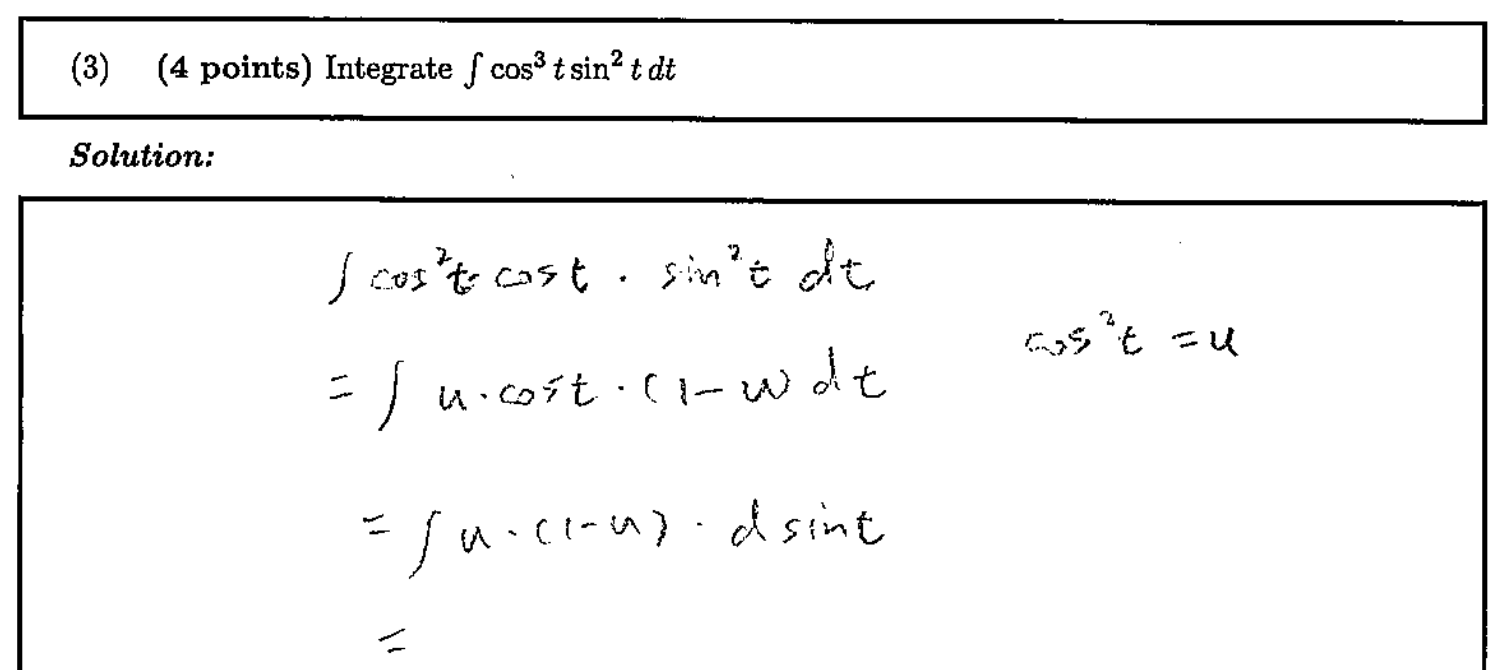}
\end{center}

{\emph {Without Prompt 1 and 2, one example of GPT-4.1-mini’s evaluation was}}: 
\begin{quote}
    The student begins by rewriting the integrand as
$\cos^2 t \cos t \sin^2 t$, which is correct. However, the substitution $u=\cos^2 t$ is not a standard or helpful substitution for this integral.
The student then attempts to express $\sin^2 t$ in terms of $u$, but the steps are not mathematically valid:
$\cos^2 t=u$ {\bf \emph{does not imply \( \sin^2 t = 1 - u \), since \( \sin^2 t = 1 - \cos^2 t \)}}, but the substitution is not handled correctly.
\end{quote}

\emph{Adding more explicit instructions  Prompt 1 and 2 in the system
prompt reduces such cases.}

\emph{More generally, this observation suggests the need for explicit consistency
checks when deploying LLM-based graders via the API. One promising direction
is to introduce a lightweight verification step—e.g., a two-stage procedure
that re-evaluates the generated explanation for internal logical
consistency—which we leave for future investigation.}

\end{example}

It is common for students to make an initial incorrect attempt, leave it on the page, and then proceed to find the correct solution. The example below illustrates the importance of Prompt 3, which ensures that such exploratory false starts are not penalized during evaluation.

\begin{example}\label{example:intialtry}
\emph{The following is a sample from Math 2B. The problem is to integrate}
\[
\int s\ln s\, ds.
\]

\begin{center}
\includegraphics[scale=0.4]{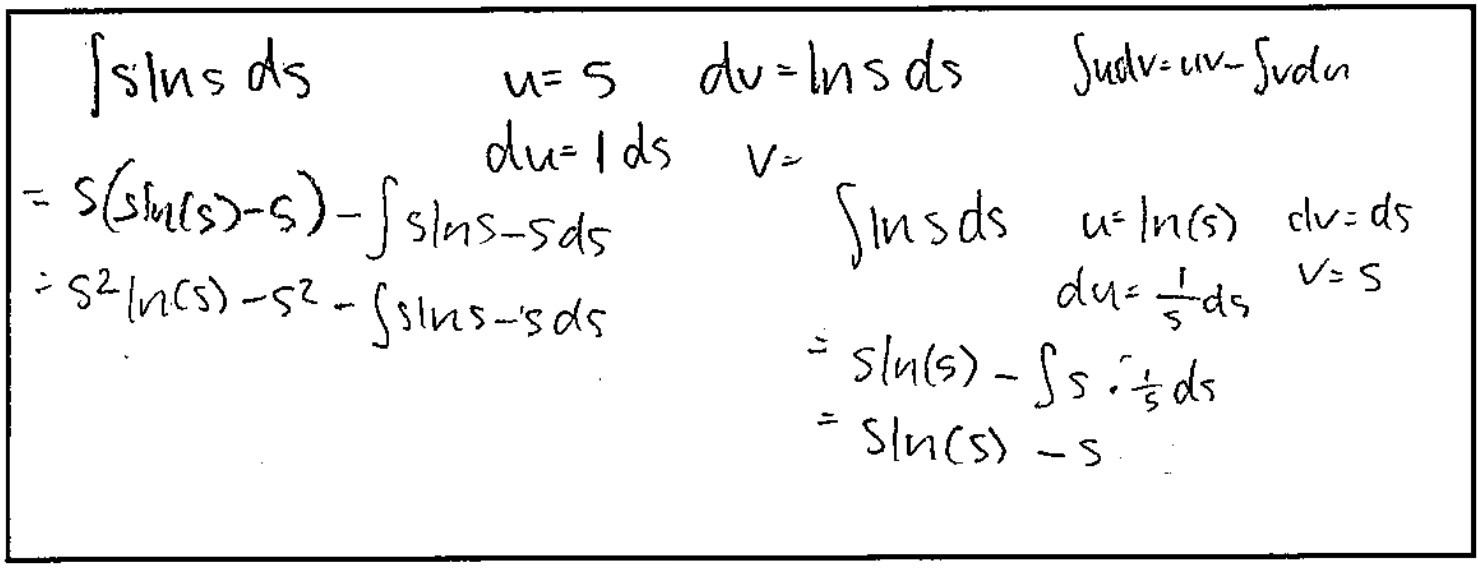}
\end{center}

\emph{As we can see, the student first makes an incorrect substitution and then corrects it later without erasing the earlier steps. Without Prompt 3, ChatGPT 4.1-mini produced the following evaluation, deducting 1 point for the initial wrong attempt (excerpted for relevance):}

\medskip
\begin{quote}
The student initially attempts integration by parts but incorrectly chooses $u = s$ and $dv = \ln(s)\, ds$, which is not a valid or helpful choice since the integral of $\ln(s)\, ds$ is nontrivial and not properly computed. However, the student later correctly restates the problem with $u = \ln(s)$ and $dv = s\, ds$, which is the correct setup and earns 1 point for correct choice and setup (.... we omit some parts here). However, because the initial part of the student's solution contains major errors and confusion in the application of integration by parts, {\bf \emph{the overall score is reduced to 2 points. The student demonstrated understanding of the correct method and obtained the correct final answer but made significant errors in the initial attempt.}}
\end{quote}
\medskip

With Prompt 3 included in the system message, this issue is resolved. Below is the revised output produced by GPT 4.1-mini:

\begin{quote}
    The student initially attempts integration by parts but makes several notation and setup errors in the first part of their solution, such as incorrectly assigning $u = s$ and $dv = \ln s\, ds$, and mixing up $du$ and $dv$. {\bf However, the student correctly abandons the incorrect start and then provides a fully correct and complete final solution} using the proper choices $u = \ln s$ and $dv = s\, ds$. The final answer matches the official correct answer exactly, including the constant of integration $C$. Since the final solution and answer are correct and logically coherent, full credit (3 points) is awarded despite the earlier false start.
\end{quote}
\end{example}

\subsection{Comparison of Two AI Grading Rubrics: Case Study}
\label{app:dual-rubrics}

In this section, we present a representative pair of rubrics (flexible vs.\ fixed) for the following quiz problem to illustrate how our rubric framework is applied in practice:

\textbf{Problem: (3 points)} Find the sum of the series
\[
\sum_{n=3}^{\infty} \left( \frac{2^{n-1}}{3^{n+3}} \right).
\]

\begin{tcolorbox}[boxrule=0.4pt,arc=2pt,left=6pt,right=6pt,top=4pt,bottom=4pt]
\textbf{Rubric 1: Flexible rubric (Mathematical Logic; accepts reindexing and both methods).}

\begin{itemize}
  \item \textbf{3.0 pts} Reasoning is coherent and the final answer simplifies to $\frac{4}{243}$ using either the direct infinite-sum formula or the partial-sum--then--limit method.
  \item \textbf{Partial credit} for any correct components: correctly identifying geometric structure/ratio; a correct first included term \emph{under the student’s indexing}; a correct partial-sum setup with $|r|<1 \Rightarrow r^N \to 0$; etc.
\end{itemize}

\paragraph{Reindexing / Equivalent Forms (Policy).}
Award full credit for any correct reindexing (e.g., $n \mapsto n+k$, changing bounds, moving constants) provided the student’s first included term and ratio are consistent. Treat algebraically equivalent forms as equal (e.g., $\frac{1}{54}\left(\frac{2}{3}\right)^n$ and $\frac{1}{81}\left(\frac{2}{3}\right)^{n-1}$). Do not penalize normalization choices; deduct only if base term or bounds are inconsistent.
\end{tcolorbox}

In other problems, we may further refine flexibility by specifying bounded score ranges for key rubric components (e.g., awarding $1$--$2$ points for establishing the validity of a critical intermediate step), rather than using only fixed point deductions as the following fixed rubric.

\begin{tcolorbox}[boxrule=0.4pt,arc=2pt,left=6pt,right=6pt,top=4pt,bottom=4pt]
\paragraph{Rubric 2: Fixed rubric (point map; explicitly handles reindexing).}

\begin{enumerate}
  \item \textbf{Geometric identification \& ratio (1.0 pt).} States/uses that the series is geometric and correctly finds $r=\frac{2}{3}$ (or rewrites to an equivalent geometric form).
  \item \textbf{First included term / correct reindexing (1.0 pt).} Computes the correct first \emph{included} term for their indexing (e.g., $a_3=\frac{4}{729}$) or uses a correct reindexing with a consistent base term.
  \item \textbf{Summation step and final value (1.0 pt).} Applies $S=\frac{a}{1-r}$ correctly or gives a correct partial-sum formula and justifies the limit via $|r|<1$. Final answer must simplify to $\frac{4}{243}$. If the setup is perfect but with a minor arithmetic slip, award $0.5/1.0$ here.
\end{enumerate}

\paragraph{Scoring Guide.}
$3.0$ --- All three components correct.\\
$2.0$--$2.5$ --- Minor algebra/indexing slip; or partial-sum path omits an explicit limit but clearly implies $r^N \to 0$.\\
$1.0$--$1.5$ --- Partial progress (e.g., found $r$ or the base term) but incorrect/incomplete summation or limit.\\
$0$--$0.5$ --- Little relevant work or wrong method (e.g., telescoping).
\end{tcolorbox}

\bigskip

\begin{example}\label{ex:dual-rubric}
\emph{We illustrate how the two rubrics evaluate the same student solution differently due to their distinct design principles. In this example, the OCR output is acceptable, and the student’s work is mathematically correct but presented in a non-canonical and somewhat informal manner. {\bf Rubric~1 (flexible, reasoning-based) assigns full credit (3.0 points, the same as TA's score), while Rubric~2 (fixed, point-map) assigns partial credit (2.0 points)}.}

\begin{center}
\includegraphics[scale=0.4]{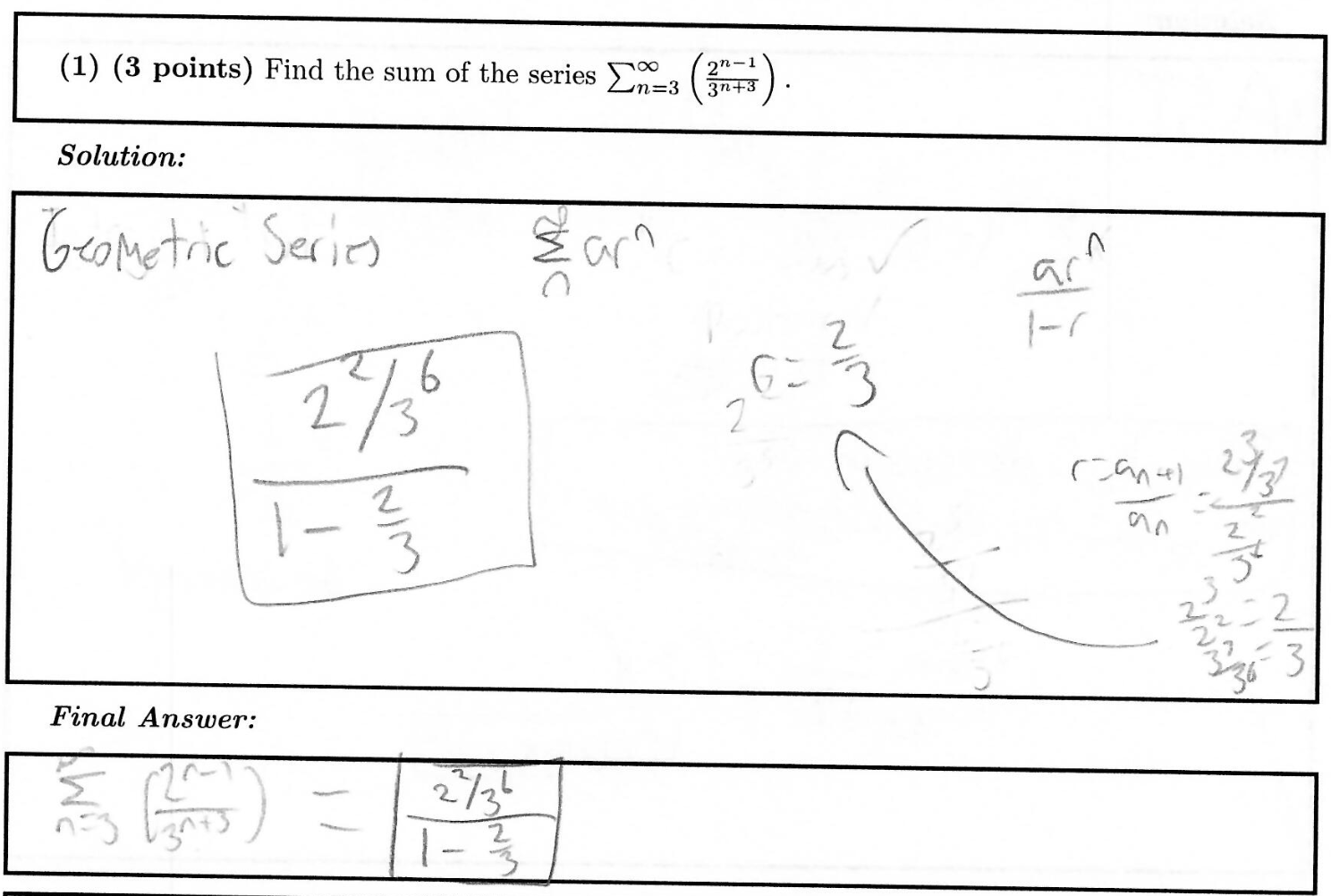}
\end{center}

{\bf \emph{Feedback from Rubric 1:}}

\begin{quote}
  The student correctly identifies the series as geometric and correctly writes the sum formula for an infinite geometric series. They correctly identify the first included term as \(\frac{2^2}{3^6}\) and the ratio as \(\frac{2}{3}\). The final answer given, \(\frac{2^2 / 3^6}{1 - \frac{2}{3}}\), is algebraically equivalent to \(\frac{4}{243}\), the correct sum. Although some intermediate steps are not fully explained or are somewhat unclear (e.g., the line with \(\frac{r - a^{n+1}}{a^n}\) is not standard notation and seems unnecessary), the key components—identifying the geometric structure, the first term, the ratio, and applying the infinite sum formula—are all present and correct. The final answer matches the reference solution in value. Therefore, full credit is awarded.
  \end{quote}  

\medskip

{\bf \emph{Feedback from Rubric 2:}}

\begin{quote}The student correctly identifies the ratio \(r = 2/3\), satisfying the first criterion (1.0 pt). However, the notation and explanation around the first term and reindexing are unclear and inconsistent. The student writes the first term as \(2^2 / 3^6 = 4/729\), which is correct, but the surrounding expressions and the fraction involving \(r\) and \(a^{n+1}\) are confusing and do not clearly establish a consistent base term or indexing. Thus, the second point for first included term / correct reindexing is only partially met, so 0.5 pt is awarded here. For the summation step, the student writes the sum as \((2^2 / 3^6) / (1 - 2/3)\), which is the correct formula for the sum of a geometric series starting at \(n=3\) with ratio \(2/3\) and first term \(4/729\). Although the student does not explicitly simplify the fraction to \(4/243\) or justify the limit of \(r^N\) to zero, the setup is essentially correct and the final boxed answer matches the correct formula. Therefore, 0.5 pt is awarded for the summation step. Overall, the solution shows understanding of the geometric series and ratio, but the explanation and notation around the first term and reindexing are incomplete and somewhat confusing, and the summation justification is minimal. Hence, total points: \(1.0 + 0.5 + 0.5 = 2.0\).
\end{quote}

\medskip

\emph{This example highlights the complementary behavior of the two rubrics. The fixed rubric assigns credit componentwise and penalizes unclear indexing and notation, whereas the flexible rubric evaluates global mathematical consistency and algebraic equivalence, resulting in full credit that aligns with the TA-assigned score. Nevertheless, we would like to emphasize that flexibility does not systematically increase scores; in other cases, the flexible rubric assigns lower scores when locally correct components are globally inconsistent.}
\end{example}

\subsection{Representative Rubric Guidance (Excerpt)}\label{app:rubircguidance}
The following excerpt illustrates rubric instructions used in our experiments to reduce unjustified point deductions due to presentation variation or OCR errors. This excerpt is representative rather than exhaustive.

\begin{quote}
\textbf{No style penalties.}{\it If the mathematics and required justification are correct, do not deduct points for presentation issues such as handwriting quality, minor typos, informal wording, or nonstandard but clear notation. If the student’s final answer appears to contain a minor typo/formatting issue or a small numerical/character discrepancy likely caused by OCR, but the correct value is clearly supported by the solution steps, do not deduct points for that discrepancy.}
\end{quote}

\subsection{AI Grading vs.\ TA Grading: A Case Study}\label{app:ai-vs-ta-case}

In this subsection, we present a representative case in which the AI-assigned score is higher than the TA-assigned score. This example illustrates how score discrepancies can arise even when both grading decisions are defensible.

\medskip

\begin{example}
\emph{The following problem is taken from a Math 2A quiz: find the local maximum values of $f$, where}
\[
f(x) = \frac{x}{x^2 + 9}.
\]
\begin{center}
\includegraphics[scale=0.25]{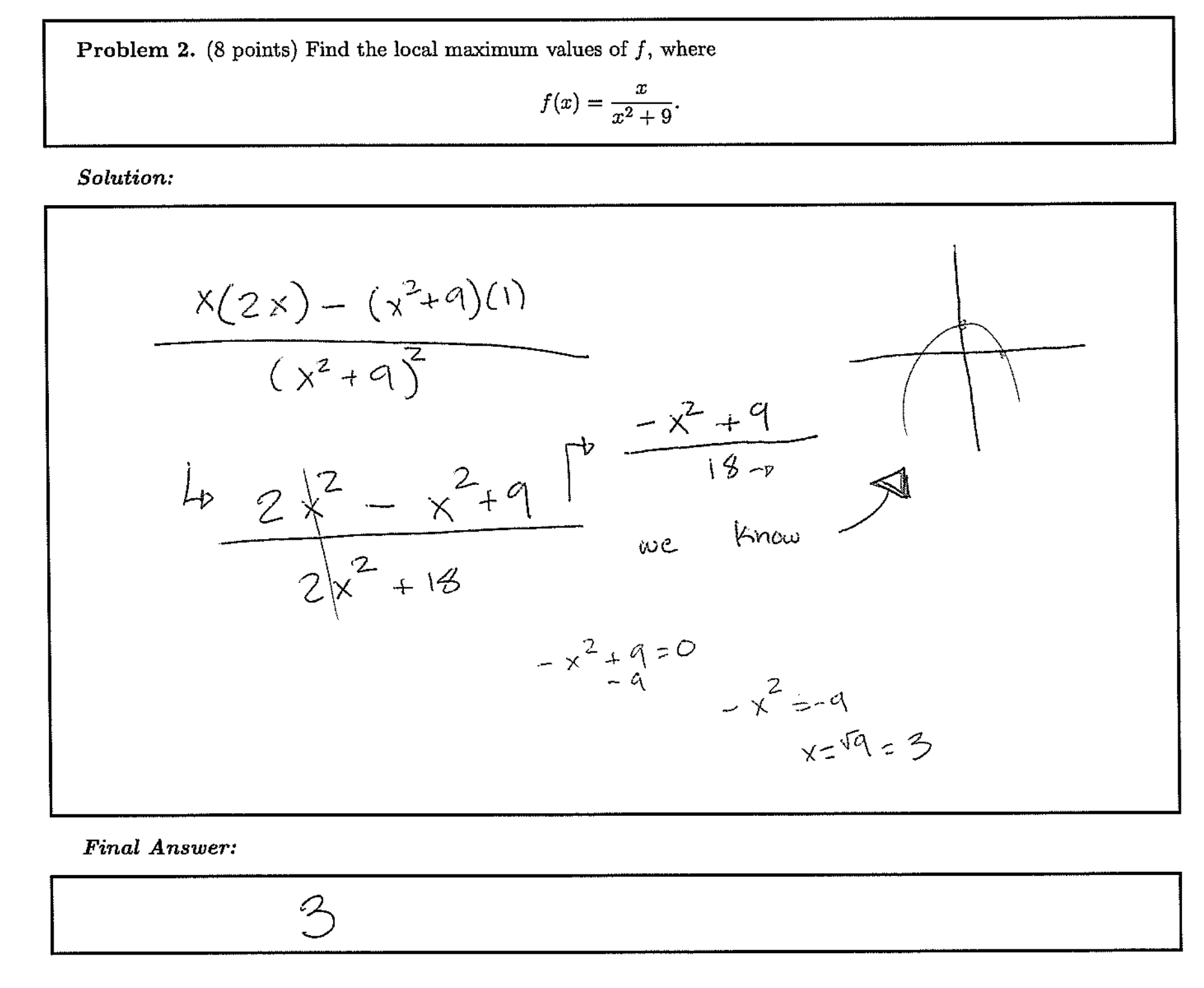}
\end{center}

\emph{For this submission, the TA assigned a score of 2 points, while the AI grader assigned a score of 5 points. The AI grader’s comments are reproduced below:}

\begin{quote}
The student attempts the quotient rule but makes algebraic errors in simplifying the derivative expression, resulting in an incorrect derivative form. This leads to a loss of 2 points on differentiation (criterion 1). The student correctly solves for $x=3$ as a critical point but misses $x=-3$, earning 1 point for finding critical points (criterion 2). There is no classification or explanation provided to confirm that $x=3$ is a local maximum, so only 1 point is awarded for identifying the max without justification (criterion 3). The student does not compute or report the maximum value $f(3)=1/6$, so 0 points are given for reporting the maximum value (criterion 4). Overall, the final answer is incomplete and lacks proper justification, but the key critical point $x=3$ is found.
\end{quote}

\emph{The AI grader applies a criterion-based rubric that separately evaluates (i) correctness of differentiation, (ii) identification of critical points, (iii) classification of extrema, and (iv) reporting of the maximum value, with partial credit awarded independently for each component. In contrast, the TA score is assigned using a coarse Gradescope rubric with a small number of score bins, shown below:}
\begin{center}
\includegraphics[scale=0.5]{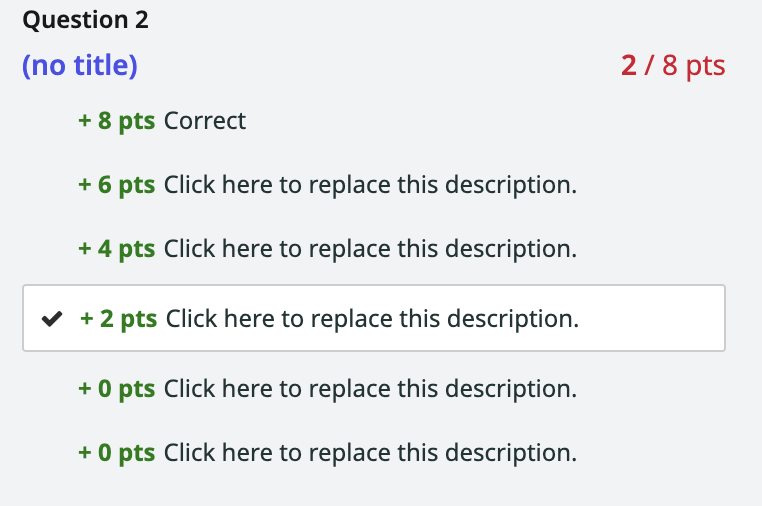}
\end{center}

\emph{This example highlights how large score differences can arise from differences in rubric structure rather than grading error. The coarse TA rubric collapses multiple aspects of partial understanding into a single low-credit category, whereas the finer-grained AI rubric aggregates partial credit across several dimensions of the solution. As a result, the AI assigns a higher total score despite the solution being clearly incomplete. Importantly, both scores are defensible under their respective rubrics. This case illustrates that AI--human score differences can reflect grading resolution and rubric expressiveness, rather than systematic bias or error by either grader.}

\end{example}
\subsection{GPT-4.1-mini vs.\ o3-mini: Comparison Results}
\label{app:4.1-o3-comparison}

Using fixed OCR outputs produced by GPT-4.1-mini, we evaluate grading behavior on a set of 57 Math 2B quizzes (171 question instances), corresponding to the same {\bf  171-solution subset} described in the main text (Section~\ref{sec:ocr-eval}). Each quiz contains the same three challenging equations. For each question, both GPT-4.1-mini and o3-mini are run three times under identical conditions. Let $\mu_{m,i}$ and $\sigma_{m,i}$ denote the mean score and run-to-run standard deviation assigned by model $m \in \{\text{4.1-mini}, \text{o3-mini}\}$ to question $i$, and define the per-question mean score difference
\[
\Delta_i = \mu_{\text{o3},i} - \mu_{\text{4.1},i}.
\]

\subsubsection{Within-model score stability}

Table~\ref{tab:within-model-stability} summarizes within-model grading stability. GPT-4.1-mini exhibits substantially lower run-to-run variability than o3-mini, as reflected by both a smaller mean standard deviation and a higher fraction of questions with zero variance. This indicates that, under low-temperature settings, GPT-4.1-mini produces more stable scores across repeated evaluations.

\begin{center}
\small
\setlength{\tabcolsep}{6pt}
\begin{tabular}{lccc}
\toprule
Model & \# Questions & Mean $\sigma$ & $\Pr(\sigma = 0)$ \\
\midrule
GPT-4.1-mini & 171 & 0.083 & 0.772 \\
o3-mini     & 171 & 0.319 & 0.725 \\
\bottomrule
\end{tabular}
\captionof{table}{Within-model grading stability across three independent runs per question, using fixed OCR inputs. $\sigma$ denotes the per-question run-to-run standard deviation, averaged across all questions.}
\label{tab:within-model-stability}
\end{center}

\subsubsection{Cross-model score differences}

Table~\ref{tab:cross-model-diff} reports aggregate statistics of cross-model score differences. On average, the mean score difference between o3-mini and GPT-4.1-mini is small, and exact agreement occurs for approximately 45\% of questions. These results suggest that, despite differences in run-to-run stability, the two models assign broadly similar scores at the aggregate level.

\begin{center}
\small
\setlength{\tabcolsep}{6pt}
\begin{tabular}{lc}
\toprule
Statistic & Value \\
\midrule
\# Questions & 171 \\
Mean $\Delta_i$ & 0.087 \\
Mean $|\Delta_i|$ & 0.315 \\
$\Pr(\Delta_i = 0)$ & 0.464 \\
\bottomrule
\end{tabular}
\captionof{table}{Cross-model score comparison between o3-mini and GPT-4.1-mini. For each question $i$, $\Delta_i$ denotes the difference in mean scores across three runs per model.}
\label{tab:cross-model-diff}
\end{center}

\subsubsection{Human reviewer comparison}\label{app:an-shiyi}

To further assess grading variability, we compare model performance using pairwise judgments from two independent human reviewers. Across both reviewers, GPT-4.1-mini and o3-mini are most often judged to perform similarly, with ties accounting for roughly 60--72\% of cases in both score proximity and feedback accuracy. When a difference is observed, o3-mini is more frequently preferred. In addition, both reviewers favor score averaging across multiple o3-mini runs. In the following two tables, Percentages are rounded to two decimal places and may not sum to exactly 100\%.

\begin{center}
\small
\setlength{\tabcolsep}{6pt}
\begin{tabular}{llr}
\toprule
\textbf{Comparison} & \textbf{Option} & \textbf{Percentage} \\
\midrule
\multirow{4}{*}{\parbox{0.44\linewidth}{\textbf{Score accuracy comparison}\\Which model is closer to the reviewer’s score?}}
& 4.1 is closer & 7.74\% \\
& o3 is closer & 29.17\% \\
& T (tie) & 63.09\% \\
\midrule
\multirow{4}{*}{\parbox{0.44\linewidth}{\textbf{Evaluation accuracy comparison}\\Which model gives more accurate feedback?}}
& 4.1 is more accurate & 3.59\% \\
& o3 is more accurate & 23.35\% \\
& T (tie) & 73.06\% \\
\midrule
\multirow{2}{*}{\parbox{0.44\linewidth}{\textbf{Averaging improves accuracy?}\\(o3 run three times; select run closest to the average score)}}
& Yes & 62.22\% \\
& No & 37.78\% \\
\bottomrule
\end{tabular}
\captionof{table}{Reviewer~1: pairwise comparison of GPT-4.1-mini (``4.1'') vs.\ o3-mini (``o3''). }
\label{tab:reviewer1-shiyi-summary}
\end{center}

\medskip

\begin{center}
\small
\setlength{\tabcolsep}{6pt}
\begin{tabular}{llr}
\toprule
\textbf{Comparison} & \textbf{Option} & \textbf{Percentage} \\
\midrule
\multirow{4}{*}{\parbox{0.44\linewidth}{\textbf{Score accuracy comparison}\\Which model is closer to the reviewer’s score?}}
& 4.1 is closer & 14.97\% \\
& o3 is closer & 17.96\% \\
& T (tie) & 67.06\% \\
\midrule
\multirow{4}{*}{\parbox{0.44\linewidth}{\textbf{Evaluation accuracy comparison}\\Which model gives more accurate feedback?}}
& 4.1 is more accurate & 9.58\% \\
& o3 is more accurate & 25.15\% \\
& T (tie) & 65.28\% \\
\midrule
\multirow{3}{*}{\parbox{0.44\linewidth}{\textbf{Averaging improves accuracy?}\\(o3 run three times; select run closest to the average score)}}
& Yes & 51.11\% \\
& No & 26.67\% \\
& Not Clear & 22.22\% \\
\bottomrule
\end{tabular}
\captionof{table}{Reviewer~2: pairwise comparison of GPT-4.1-mini (``4.1'') vs.\ o3-mini (``o3''). }
\label{tab:reviewer2-an-summary}
\end{center}

\paragraph{Relation to majority-vote self-consistency \cite{wang2023}. }
Our multi-run stabilization selects the run whose score is closest to the mean across three runs, rather than the most frequent score as in majority-vote self-consistency. To examine the practical difference, we analyzed the 45 cases in which the three o3-mini runs produced non-identical scores. In the majority of these cases, two runs shared the same score, making majority voting and closest-to-mean selection equivalent. Only five cases exhibited three distinct scores; among these, outcomes were mixed across reviewers, with no consistent preference for either rule. This suggests that, at the course scale and with three runs, closest-to-mean selection behaves similarly to majority voting in most cases, while differing only in a small number of genuinely ambiguous instances. We restrict to three runs to balance stability and cost in course-scale deployment, where majority voting over many samples would be impractical. In the rare cases where all three runs disagree, the discrepancy reflects intrinsic grading ambiguity and such cases are treated conservatively (e.g., flagged for downstream review).As a next step, we will quantify how often three-way disagreements occur at scale; if they remain rare, a conservative policy is to defer such cases to human review, whereas more frequent occurrences would motivate additional stabilization strategies that jointly consider both scores and feedback, rather than score aggregation alone.

\subsubsection{Inter-reviewer variability}\label{app:tworeviewerdifference}

Finally, to contextualize AI--human score differences, we examine inter-reviewer variability among human graders using the same criterion-based rubric as the AI grader. Two independent reviewers graded the same 57 quizzes (171 question instances). Figure~\ref{fig:two-reviewer-diff} shows the distribution of per-question score differences. While exact agreement occurs for a substantial fraction of questions (approximately 45\%), nontrivial discrepancies spanning multiple score levels are common. This illustrates the inherent subjectivity of human grading even under a shared, well-specified rubric and provides a natural baseline for interpreting AI--human score differences.

\begin{center}
\includegraphics[scale=0.5]{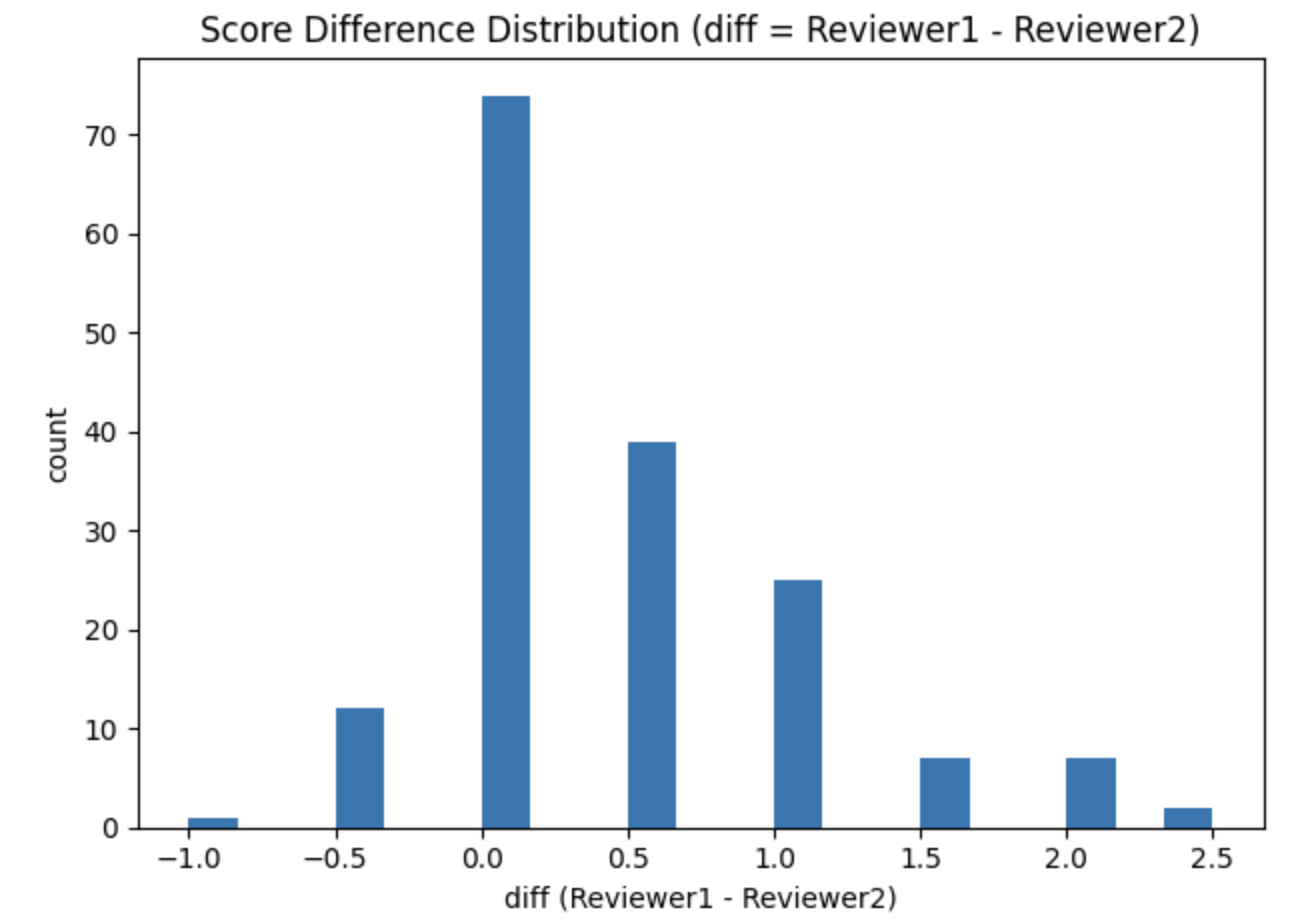}
\captionof{figure}{Distribution of per-question score differences between two human reviewers (Reviewer~1 minus Reviewer~2).}
\label{fig:two-reviewer-diff}
\end{center}

\subsection{Example Student-Facing Message}\label{app:messagetostudent}
Below is an anonymized example of a message sent to a student in Math 2B, including both the assigned score and the accompanying feedback. All messages followed the same format; only the per-question content differed.

\begin{quote}
    
Dear (student name),

\medskip

Thank you for your participation in the AI grading tests. Your test AI grading result is as follows:

\medskip

Question 1:

Points: 5

Evaluation: The student's work is correct. They correctly applied integration by parts, correctly computed the antiderivative, and accurately evaluated the limits. The final expression, $(5/9)e^6 + 1/9$, is algebraically equivalent to the expected answer $1/9(5e^6 + 1)$. There are no errors in the computation or algebra, so full credit is awarded.

\medskip

Question 2:

Points: 2

Evaluation: You correctly recalled the identity $sec^2x = 1 + tan^2x$ and attempted to rewrite the integrand accordingly, which is a good start (criterion 1). However, the error occurs when you express $sec^6x$ as $(sec^2x)^3$ and then replace each $sec^2x$ by $1 + tan^2x$. The proper approach is to factor $sec^6x$ as $sec^4x \cdot sec^2x$ so that one can use the substitution $u = tan x$ (with $du = sec^2x dx$) and rewrite $sec^4x$ as $(1 + tan^2x)^2$ rather than $(1 + tan^2x)^3$. This misstep in rewriting the integrand leads to an incorrect polynomial (criterion 3) and, consequently, an incorrect antiderivative (criteria 4 and 5). While you did integrate the resulting terms correctly, the algebraic setup was flawed. For these reasons, you earn 2 out of 5 points.

\medskip

Total Score: 7

\medskip

Please note: AI grading is for reference only. Scores assigned by your TA in Gradescope are the official scores.
Our AI grading has two LLM-based steps—(1) OCR to read handwritten work and (2) evaluation of the OCR text. Because LLMs are probabilistic, occasional mistakes can occur.
\end{quote}

\end{document}